\documentclass[runningheads]{llncs}

\usepackage{accv}

\usepackage[T1]{fontenc}

\usepackage{accvabbrv}

\usepackage{graphicx}
\usepackage{booktabs}
\usepackage{amsmath}

\usepackage[accsupp]{axessibility}  

\usepackage{hyperref}

\usepackage{orcidlink}

\usepackage{xspace}

\newcommand{\aka}{\emph{a.k.a.}\xspace}
\newcommand\carlaParam[1]{\texttt{#1}\xspace}%

\newcommand{\paramCloudiness}{\carlaParam{cloudiness}}
\newcommand{\paramFogDensity}{\carlaParam{fog density}}
\newcommand{\paramPrecipitation}{\carlaParam{precipitation}}
\newcommand{\paramPrecipitationDeposits}{\carlaParam{precipitation deposits}}
\newcommand{\paramSunAltitudeAngle}{\carlaParam{sun altitude angle}}
\newcommand{\paramSunAzimuthAngle}{\carlaParam{sun azimuth angle}}
\newcommand{\paramWindIntensity}{\carlaParam{wind intensity}}

\newcommand{\LS}{\emph{LS}\xspace}
\newcommand{\TS}{\emph{TS}\xspace}

\global\long\def\realNumbers{\mathbb{R}}%
\global\long\def\sampleSpace{\Omega}%
\global\long\def\eventSpace{\Sigma}%
\global\long\def\measurableSpace{(\sampleSpace,\eventSpace)}%
\global\long\def\allDomains{\mathbb{D}_{\measurableSpace}}%
\global\long\def\domain{d}%
\global\long\def\domainRef{d_R}%
\global\long\def\domainTarget{d_T}%
\newcommand\domainSource[1][1=i]{d_{S_{#1}}}%
\global\long\def\numSources{n_S}%
\global\long\def\domainInODD{d_{\in\odd}}%
\global\long\def\domainOutODD{d_{\notin\odd}}%
\global\long\def\probabilityMeasure{P}%
\global\long\def\allImages{\mathbb{I}}%
\global\long\def\anImage{i}%
\global\long\def\rvImage{I}%
\global\long\def\allWeathers{\mathbb{W}}%
\global\long\def\aWeather{w}%
\global\long\def\rvWeather{W}%
\global\long\def\numWeathers{n_W}%
\global\long\def\aBag{b}%
\global\long\def\rvBag{B}%
\global\long\def\weight{\omega}%
\global\long\def\superiorityGT{\pi}

\newcommand\angledegrees[1]{#1^{\circ}}%
\global\long\def\credibleLevel{l}%

\newcommand{\carla}{\emph{CARLA}\xspace}
\newcommand{\lampe}{\emph{LAMPE}\xspace}
\newcommand{\zuko}{\emph{ZUKO}\xspace}
\newcommand{\cvxpy}{\emph{CVXPY}\xspace}

\newcommand{\mda}{\emph{MDA}\xspace}
\newcommand{\odd}{\emph{ODD}\xspace}
\newcommand{\sbi}{\emph{SBI}\xspace}

\newcommand{\nsf}{\emph{NSF}\xspace}
\newcommand{\npe}{\emph{NPE}\xspace}
\newcommand{\cnn}{\emph{CNN}\xspace}
\newcommand{\ppc}{\emph{PPC}\xspace}

\newcommand{\pdf}{\emph{PDF}\xspace}
\newcommand{\vtov}{\emph{V2V}\xspace}

\newcommand{\resnet}{\emph{ResNet-50}\xspace}
\newcommand{\dino}{\emph{DINOv2}\xspace}
\newcommand{\clip}{\emph{CLIP}\xspace}

\makeatletter
\newcommand{\printfnsymbol}[1]{%
  \textsuperscript{\@fnsymbol{#1}}%
}
\makeatother

\begin{document}

\title{Physically Interpretable\\ Probabilistic Domain Characterization} 


\author{
Anaïs Halin\thanks{Equal contributions. \email{\{anais.halin,s.pierard\}@uliege.be}}\inst{1}\orcidlink{0000-0003-3743-2969}
\and Sébastien Pi\'erard\printfnsymbol{1}\inst{1}\orcidlink{0000-0001-8076-1157}
\and Renaud Vandeghen\inst{1}\orcidlink{0009-0003-1752-1195}
\and\\ Benoît G\'erin\inst{2}\orcidlink{0009-0007-3249-0656}
\and Maxime Zanella\inst{2,4}\orcidlink{0009-0009-4030-3704}
\and Martin Colot\inst{3}\orcidlink{0009-0000-5852-573X}
\and Jan Held\inst{1}\orcidlink{0009-0005-7907-2895}
\and\\ Anthony Cioppa\inst{1}\orcidlink{0000-0002-5314-901}
\and Emmanuel Jean\inst{5}\orcidlink{0009-0004-1544-7080}
\and Gianluca Bontempi\inst{3}\orcidlink{0000-0001-8621-316X}
\and\\ Saïd Mahmoudi\inst{4}
\and Benoît Macq\inst{2}\orcidlink{0000-0002-7243-4778}
\and Marc Van Droogenbroeck\inst{1}\orcidlink{0000-0001-6260-6487}
}

\authorrunning{A.~Halin, S. Piérard, \etal}

\institute{
Montefiore Institute, University of Li\`ege (ULiège), Li\`ege, Belgium
\and Catholic University of Louvain (UCLouvain), Louvain-la-Neuve, Belgium
\and Universit\'e Libre de Bruxelles (ULB), Brussels, Belgium
\and University of Mons (UMons), Mons, Belgium
\and Multitel research \& innovation centre, Mons, Belgium
}

\maketitle

\begin{abstract}
Characterizing domains is essential for models analyzing dynamic environments, as it allows them to adapt to evolving conditions or to hand the task over to backup systems when facing conditions outside their operational domain. 
Existing solutions typically characterize a domain by solving a regression or classification problem, which limits their applicability as they only provide a limited summarized description of the domain. 
In this paper, we present a novel approach to domain characterization by characterizing domains as probability distributions.
Particularly, we develop a method to predict the likelihood of different weather conditions from images captured by vehicle-mounted cameras by estimating distributions of physical parameters using normalizing flows.
To validate our proposed approach, we conduct experiments within the context of autonomous vehicles, focusing on predicting the distribution of weather parameters to characterize the operational domain. This domain is characterized by physical parameters (absolute characterization) and arbitrarily predefined domains (relative characterization).
Finally, we evaluate whether a system can safely operate in a target domain by comparing it to multiple source domains where safety has already been established.
This approach holds significant potential, as accurate weather prediction and effective domain adaptation are crucial for autonomous systems to adjust to dynamic environmental conditions.
\end{abstract}

\section{Introduction}
\label{sec:intro}
Advances in computer vision allow widespread camera monitoring, but diverse weather conditions lead to visually different data, sometimes impacting the performance of high-level tasks like object detection and surveillance. Current solutions often lack generalizability across multi-weather scenarios, highlighting the need for adaptive methods that can process visual data under diverse conditions.

More specifically, weather conditions significantly affect the perception capabilities of autonomous driving systems~\cite{perrels2015weather}, particularly under harsh conditions, such as heavy rain or fog, which can compromise their ability to operate safely.
Therefore, it is essential to develop reliable approaches to analyze the environment, irrespective of the weather conditions. 
Detecting critical circumstances such as extreme weather events allow the system to respond appropriately within and outside its \emph{Operational Design Domain} (\odd), defined by SAE International~\cite{Sae2021Taxonomy} as the ``operating conditions under which a given driving automation system, or feature thereof, is specifically designed to function, including, but not limited to, environmental, geographical, and time-of-day restrictions, and/or the requisite presence or absence of certain traffic or roadway characteristics''. 
 
\begin{figure}[t!]
  \centering
    \includegraphics[width=\linewidth]{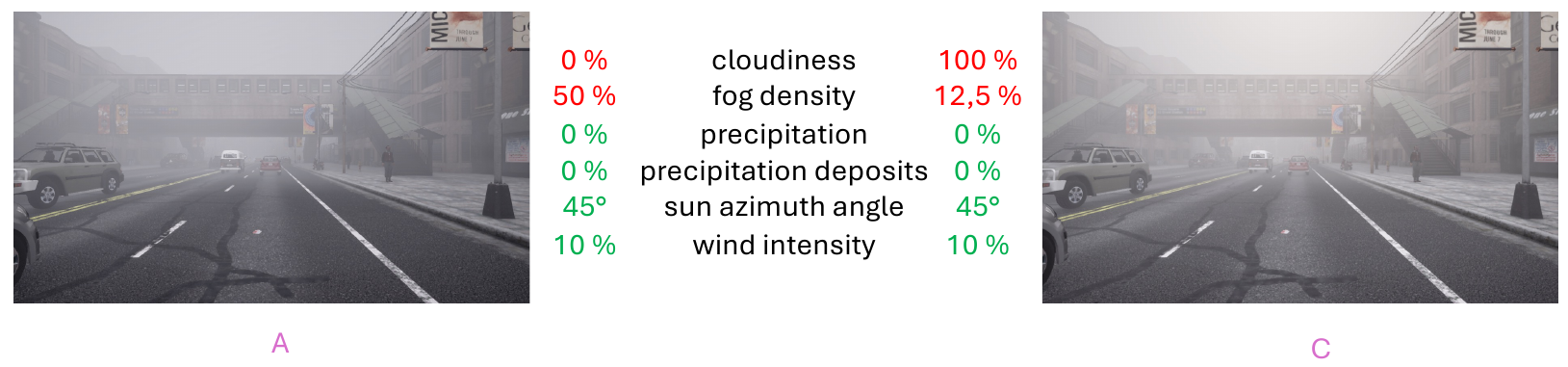}
  
  \caption{
  The two synthetic images, A and C, generated using the \carla software, appear almost identical, despite being acquired under very different weather conditions. 
  This highlights the challenge of information loss from sensors like cameras when dealing with weather-related physical parameters. As a result, predictions that diverge from the ground truth in such ambiguous cases should not be penalized during evaluation.} 
    \label{fig:observation-1-ambiguous-cases} 
\end{figure}

However, predicting exact weather conditions is challenging due to the many ambiguous cases.
A single observation of the environment (\eg, recorded by a vehicle-mounted camera) could be characterized by multiple sets of weather parameters. 
Fig.~\ref{fig:observation-1-ambiguous-cases} presents two synthetic images, generated using the \carla software~\cite{Dosovitskiy2017CARLA}, that appear nearly identical but were captured under different weather conditions. This shows the ambiguities inherent to characterizing the domain based on a single observation (\ie, image).

\begin{figure}[t!]
  \centering
    \includegraphics[width=\linewidth]{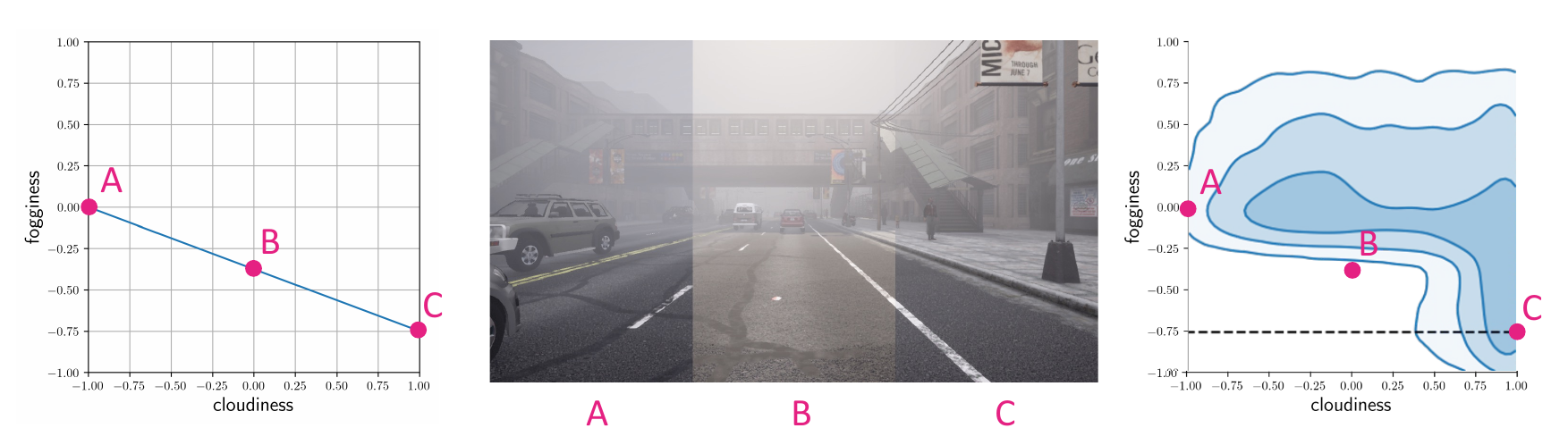}
  
  \caption{
  Considering the images A and C from Fig.~\ref{fig:observation-1-ambiguous-cases}, generating an image B for the arithmetically averaged parameters, leads to an image very different from A and C.
  In other words, there are images $\anImage$ such that the probability $\probabilityMeasure(\rvImage=\anImage,\rvWeather=\hat{\aWeather}(\anImage))$ of the pair (image, estimated weather) is zero when the estimated weather is the expected value of the weather knowing the image, $\hat{\aWeather}(\anImage)=E[\rvWeather|\rvImage=\anImage]$, as done in regression. Working with distributions of weather parameters (as shown using a contour plot on the right-hand side) as proposed in this work, rather than predicting specific values for each parameter, \ie, regression (as shown on the left-hand side), avoids this problem.} 
    \label{fig:reg_vs_normalizing_flow} 
\end{figure}

Current methods predict weather conditions through regression or classification~\cite{ibrahim2019weathernet, Introvigne2024Realtime, li2017multi}. 
However, those approaches only produce a single crisp answer, lacking the insight about the ambiguities of characterizing the weather conditions. 
Fig.~\ref{fig:reg_vs_normalizing_flow} illustrates how the two scenarios from Fig.~\ref{fig:observation-1-ambiguous-cases}, when processed with an intermediate set of values, result in very different visual representations. 

 In this work, we propose a novel solution based on a probabilistic characterization of the weather.
Historically, statistical inference relying on likelihood estimation was intractable when dealing with high dimensional data~\cite{Cranmer2020TheFrontier}. Recently, one class of density estimation techniques based on neural networks called normalizing flows~\cite{9089305, JMLR:v22:19-1028, Rozet2022Zuko} has become quite popular to solve this kind of problem. These models learn invertible transformations to go from complex distributions to more handy ones, and can therefore be used for modeling weather parameters from highly complex image and weather distributions.

These novel techniques allow us to express the domain of an image by a probabilistic distribution, \eg, weather conditions, compared to current deterministic approaches.  
More specifically, we show on three consecutive tasks how various weather conditions can be predicted based (1)~on a single color image acquired in front of a vehicle, (2)~on a bag of color images (\emph{absolute characterization}), and (3)~how the current domain is related to arbitrarily chosen source domains (\emph{relative characterization}).

Notions similar to \odd are very common in other fields where equivalent terms are used, such as \emph{Operational Envelope} for maritime and \emph{Operational Context} for railroad~\cite{tonk2022operational}. Additionally, other fields are facing domain shifts, such as in medical imagery from different acquisition devices~\cite{gerin2024exploring}. We argue that our approach could serve a large range of practical applications. 
For example, determining the most suitable model for a given scenario within a fleet of lightweight AI models able to analyze the environment, as proposed in previous works~\cite{Mansour2008Domain, Pierard2023Mixture, Gerin2024MultiStream}, detecting significant domain transitions to collect new data for domain adaptation methods that rely on buffers or adaptable internal statistics~\cite{Wang2020Tent-arxiv, wang2022continual, Gong2022NOTE, yuan2023robust, Wang2023Dynamically, Houyon2023Online} or activating adaptation mechanisms based on clustering~\cite{Tang2020Unsupervised, Boudiaf2022Parameterfree, zanella2024test, zanella2024boosting}. 

We summarize our contributions as follows:
(1)~We propose a novel probabilistic methodology to characterize domains in the case of autonomous vehicles driving in various weather conditions.  
(2)~We demonstrate that simulation-based inference (normalizing flows) is adequate to obtain distribution for weather parameters that are used to characterize the domain and compare different backbones for features extraction.
(3)~Based on this weather domain characterization, we show how to characterize a new target domain as a mixture of source models.

\section{Related Work}
\label{sec:related-work}

\subsection{Predicting Weather Conditions from Images}
Prediction of weather conditions from images was first formulated as a single-label classification task (\eg, sunny, cloudy, or foggy). In 2014, Lu~\etal~\cite{Lu2017TwoClass} proposed a binary classification task between sunny and cloudy weathers, using features extracted from visual cues such as the sky, shadows, reflections, contrast, and haze. Later, Guerra~\etal~\cite{Guerra2018Weather} proposed a multi-class dataset extending the scope of the classification task to rain, snow, and fog.
Recent works focus on optimizing \emph{Convolutional Neural Network} (\cnn) to obtain strong features for common and uncommon weather conditions~\cite{li2020multi, zhang2021multi}. However, weather conditions can hardly be represented by crisp classification due to its continuous nature.

To reach a more realistic description of intricate relations between weather conditions, recent methods simultaneously regress several physical parameters~\cite{ibrahim2019weathernet, Introvigne2024Realtime}. To increase interpretability, Li~\etal~\cite{li2017multi} also assign cues of weather characteristics to each pixel. However, these methods still lack the ability to represent ambiguous scenarios (see Fig.~\ref{fig:reg_vs_normalizing_flow}).
In this work, we predict the joint distribution of weather parameters by proposing a novel method based on normalizing flows.

\subsection{Handling Weather Conditions for Autonomous Driving}
Autonomous car driving systems need to be efficient under all weather conditions.
Some methods propose to integrate a generalization step to the model to remove the environmental influences on the acquired images, by introducing a style layer inspired by images style-transfer neural networks~\cite{rebut2021styleless} or through adversarial training~\cite{li2023domain}. 
Generalized features obtained from alignment of source and target domains tend to be suboptimal, as they do not consider the task. To improve the estimation, a solution is to add a domain adaptation step~\cite{lee2019drop}.
Jeon~\etal~\cite{jeon2024raw} further improved their estimation by adding an unsupervised domain adaptation step after domain alignment.

Many studies have also highlighted the importance of a strong \odd definition to properly assess the ability of automatic driving systems to work in given conditions regarding weather, location, other vehicles on the road, state of the car sensors and many other environmental parameters~\cite{PAPPALARDO2022631, Gyllenhammar2020Towards, Colwell2018AnAutomated}. Many strategies have been proposed to evaluate different situations according to specific evaluations of potential damage cost~\cite{Sun2022Acclimatizing, lee2020identifying} and define the boundaries of the \odd.
In this work, we characterize the domain by a probability distribution focusing on weather parameters in autonomous driving environment.

\subsection{Probabilistic Modeling of Parameters Distributions}
We propose to leverage recent observations in the \emph{Simulation-Based Inference} (\sbi) literature~\cite{Cranmer2020TheFrontier, Vasist2023Neural} for domain characterization. This literature has seen a rapid expansion thanks to new density estimation techniques in problems where likelihood estimation was often intractable, especially for high dimensional data. One class of these density estimation techniques based on neural networks is normalizing flows~\cite{JMLR:v22:19-1028}. 
The principle consists in transforming an arbitrarily chosen distribution (\eg, a Gaussian) into the desired distribution. Different types exist, such as the \emph{Neural Spline Flow} (\nsf) type~\cite{durkan2019neural} that can be used in two ways: (1)~to determine the value of the \emph{Probability Density Function} (\pdf) at a given point and (2)~to draw samples at random. Different techniques can be used to learn them, \eg the \emph{Neural Posterior Estimation} (\npe) technique~\cite{Lueckman2017Flexible, Greenberg2019Automatic}.

There are a few techniques to analyze the performance of models predicting distributions. \emph{Coverage Plots}~\cite{hermans2021trust} show, objectively and quantitatively, whether the distribution prediction models are underconfident (\ie, conservative), calibrated, or overconfident. Another technique, specific for parameter distributions (\eg, weather parameters) predicted from an observation (\eg, an image) and widespread in the field of \sbi, is known as \emph{Posterior Predictive Check} (\ppc)~\cite{robertson2014assessing}. It consists in drawing parameters at random from a predicted distribution, injecting these parameters into a simulator or physical system and comparing the resulting observations with the one from which the distribution was predicted. In this work, we leverage those analysis techniques for assessing the quality of our weather characterization models.

\section{The Three Fundamental Tasks Behind the Physically Interpretable
Probabilistic Domain Characterization}
\label{sec:experiments}
Our experiments are organized around three different tasks, involving a prediction of the distribution of weather conditions given some images acquired by color cameras placed in front of vehicles. Before elaborating on these tasks in Sec.~\ref{sec:task-1}, \ref{sec:task-2}, and~\ref{sec:task-3}, we briefly introduce our framework in Sec.~\ref{sec:framework}.

\subsection{Framework}
\label{sec:framework}
\subsubsection{Mathematical Modeling.}
We denote the set of all possible values for the physical parameters (\eg weather conditions) by $\allWeathers$ and the set of all observations of interest (\eg images) by $\allImages$. We adopt the probability theory of Kolmogorov~\cite{Kolmogorov1933Grundbegriffe,Kolmogorov1950Foundations} and consider a measurable space $\measurableSpace$ as well as the (generalized) random variables $\rvWeather:\sampleSpace\rightarrow\allWeathers$ for the physical parameters and $\rvImage:\sampleSpace\rightarrow\allImages$ for the observation. Following the mathematical modeling introduced in~\cite{Pierard2023Mixture}, we consider the set $\allDomains$ of domains $\domain$ in which there is a probability measure $\probabilityMeasure_{\domain}$ on $\measurableSpace$. We see the \odd of a given autonomous system as the set of domains in which it can be used safely, no matter if this has been established by design or by testing. Thus, $\odd \subseteq \allDomains$.

\subsubsection{Data.}
\begin{table}[t]
\caption{In \carla, the weather is controlled through $13$ physical parameters. This table shows the range of values that we consider in our experiments and indicates, for each of them, if they are considered in our predictions.\label{tbl:weather-parameters}}

\begin{centering}
\resizebox{0.85\textwidth}{!}{ 
\begin{tabular}{|c|c|c|}
\hline 
parameter & range & predicted\tabularnewline
\hline 
\hline 
\paramCloudiness & $0$ to $100\%$ & yes \tabularnewline
\hline 
\paramFogDensity & $0$ to $100\%$ & yes \tabularnewline
\hline 
\paramPrecipitation & $0$ to $100\%$ & yes \tabularnewline
\hline 
\paramSunAzimuthAngle & $\angledegrees{0}$ to $\angledegrees{360}$ & no \tabularnewline
\hline 
\paramSunAltitudeAngle & $\angledegrees{-90}$ to $\angledegrees{90}$ & yes \tabularnewline
\hline 
\paramWindIntensity & $0$ to $100\%$ & yes \tabularnewline
\hline 
\paramPrecipitationDeposits & $0$ to $100\%$ & yes \tabularnewline
\hline 
\end{tabular} %
\hspace{5mm}
\begin{tabular}{|c|c|c|}
\hline 
parameter & range & predicted\tabularnewline
\hline 
\hline 
\carlaParam{fog distance} & fixed to $0.75$ & no \tabularnewline
\hline 
\carlaParam{fog falloff} & fixed to $0.1$ & no \tabularnewline
\hline 
\carlaParam{mie scattering scale} & fixed to $0.03$ & no \tabularnewline
\hline 
\carlaParam{rayleigh scattering scale} & fixed to $0.033$ & no \tabularnewline
\hline 
\carlaParam{scattering intensity} & fixed to $1.0$ & no \tabularnewline
\hline 
\carlaParam{wetness} & fixed to $0.0$ & no \tabularnewline
\hline 
\multicolumn{1}{c}{} & \multicolumn{1}{c}{} & \multicolumn{1}{c}{}\tabularnewline
\end{tabular} %
}
\par\end{centering}
\end{table}

All our experiments are performed on data generated using the \carla software, an open-source simulator for autonomous driving research~\cite{Dosovitskiy2017CARLA}. All images are acquired by a simulated camera placed in front of a vehicle called \emph{ego vehicle}. The weather is controlled through $13$ real-valued parameters (see Table~\ref{tbl:weather-parameters}). 
There are $6$ parameters for which we make predictions. Thus, $\allWeathers\subseteq\realNumbers^6$.

\subsection{Task I: Predicting Distributions of Physical Parameters}
\label{sec:task-1}
\begin{figure}[b!]
  \centering
    \includegraphics[width=\linewidth]{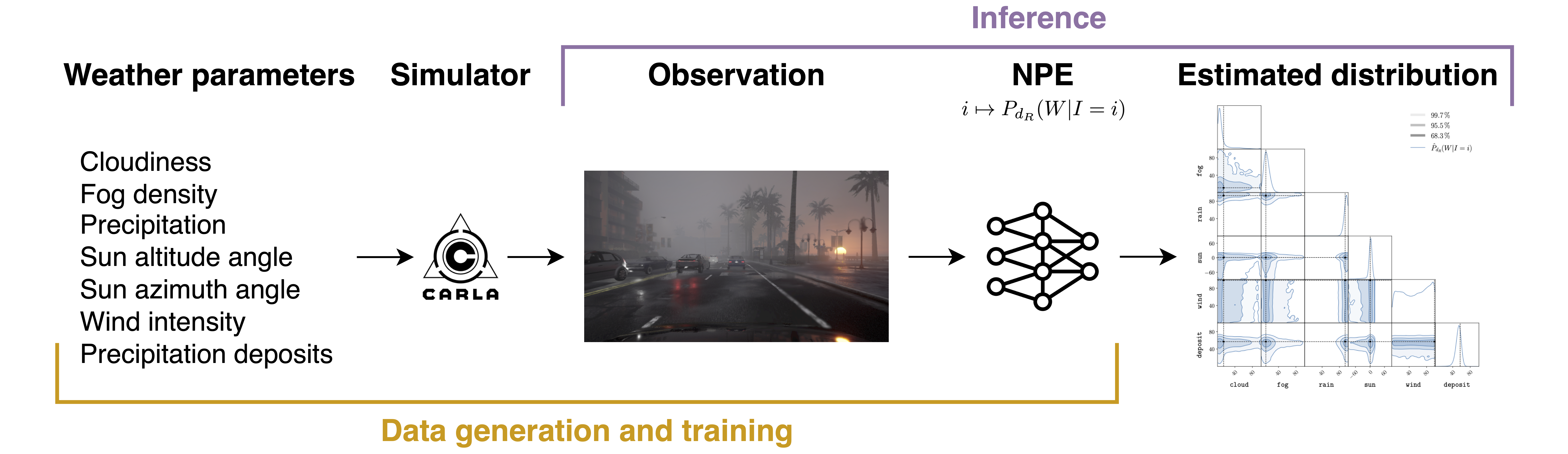}
  
  \caption{
  Task I.
  The aim of the first task is, based on an image, to predict the joint distribution of the weather parameters. For this purpose, (1)~we generate data using the \carla software for uniformly distributed weather parameters (offline), (2)~we train a \npe model using the learning set (\LS) of our generated data (offline), and (3)~we infer, given an image from the test set (\TS), the estimated weather distribution (normalizing flow) and show the result on a corner plot.
  }\label{fig:task-I} 
\end{figure}

The first task (see Fig.~\ref{fig:task-I}) consists in predicting how likely various physical parameters are, jointly. As these parameters cannot, in general, be measured directly, it is necessary to estimate the likelihood, \ie the distribution of plausible values, based on an indirect observation.

\subsubsection{Case Study.}
We study here the particular case in which the physical parameters are relative to the weather conditions and the observations are color images $\anImage\in\allImages$ acquired by cameras placed in front of vehicles. We aim at learning, offline, a deep learning model 
$
    \anImage \mapsto \hat{\probabilityMeasure}_{\domainRef}(\rvWeather\vert \rvImage=\anImage)
$, 
where $\domainRef$ denotes a domain of reference in which the probability measure  on $\measurableSpace$ is $\probabilityMeasure_{\domainRef}$. This domain is arbitrarily chosen in such a way that $\probabilityMeasure_{\domainRef}$ has a large support.

\subsubsection{Data.}
\begin{figure}[t!]
  \centering
    \includegraphics[width=\linewidth]{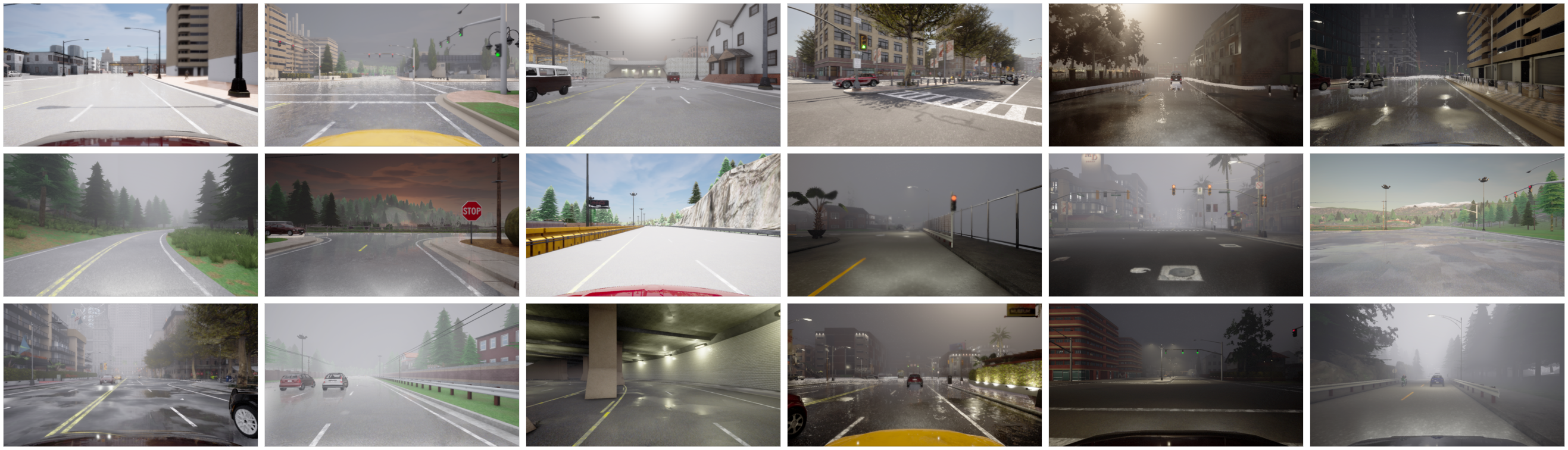}

  \caption{ 
  Excerpt of the images in our dataset generated with the \carla simulator. The ground-truth weather parameters are drawn at random for each image, following a uniform distribution with the bounds given in Table~\ref{tbl:weather-parameters}.} \label{fig:dataset-I} 
\end{figure}

We use \carla to generate a dataset with $635$k images and the corresponding ground-truth values for the weather parameters. The dataset (Fig.~\ref{fig:dataset-I}) is split into a learning set (\LS) with $600$k images ($500$k for the training set and $100$k for the validation set) and a test set (\TS) with $35$k images. Letting the model of the ego vehicle, the map, the number of pedestrians, and the number of vehicles vary brings a touch of diversity in our data. 

\subsubsection{Method.}
We consider three different (frozen) backbones to extract features from the input images: \resnet~\cite{He2016DeepResidual}, \dino~\cite{Oquab2023DINOv2-arxiv}, and \clip~\cite{Radford2021Learning}. We use the libraries \lampe~\cite{Rozet2021LAMPE} and \zuko~\cite{Rozet2022Zuko} to learn a model (of type \npe) and to manipulate the normalizing flows (of type \nsf), respectively. All predicted weather distributions are posteriors relative to the weather priors in the \LS. 
Note that \lampe and \zuko were not developed for domain characterization, but rather for \sbi. Also, to the best of our knowledge, these libraries have only been used once with very high dimensional input data~\cite{Vasist2023Neural}.  

\subsubsection{Evaluation and Results.}
Five different analyses are carried out.
\begin{figure}[p]
\begin{centering}
\subfloat[An input image $\anImage$ and the corresponding ground-truth values.\label{fig:task-I-input}]{
\begin{centering}
\noindent\begin{minipage}[t]{1\textwidth}%
\begin{center}
\scriptsize
~\hfill{}\includegraphics[height=8\baselineskip]{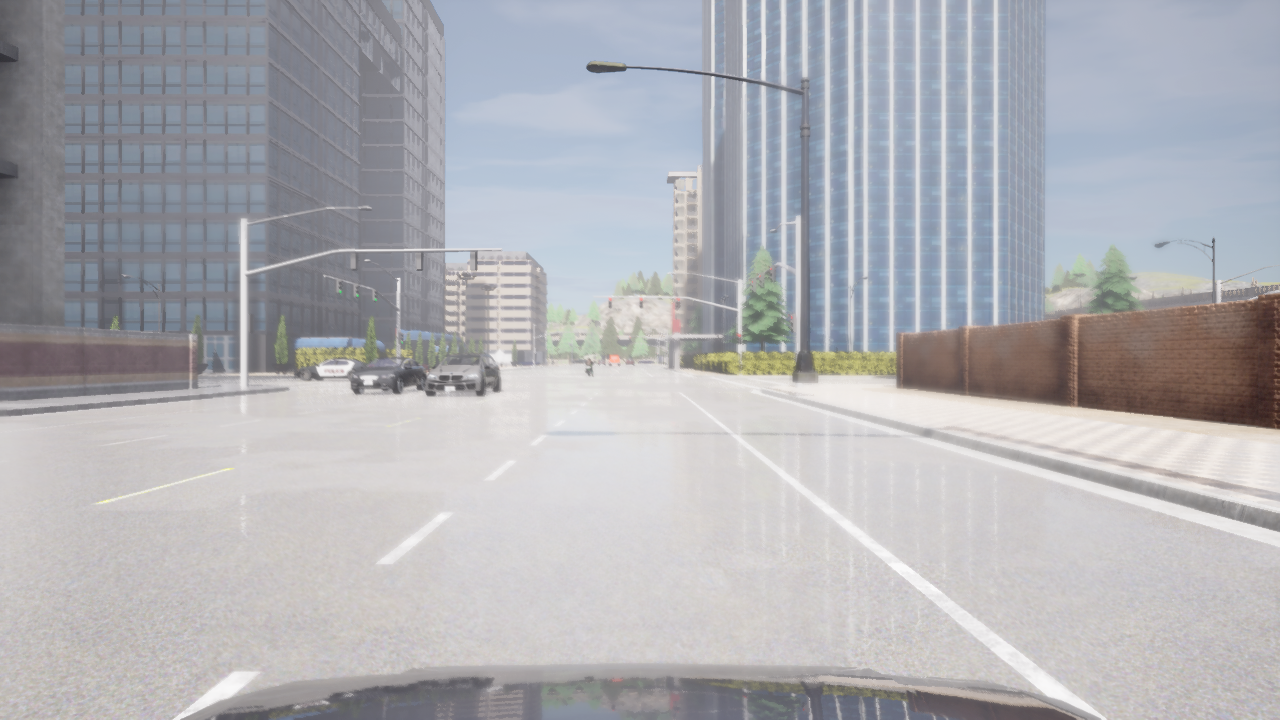}\hfill{}\hfill{}%
\begin{tabular}[b]{|c|c|}
\hline 
weather parameter & ground-truth value\tabularnewline
\hline 
\hline 
\paramCloudiness (\texttt{cloud}) & $19$ \tabularnewline
\hline 
\paramFogDensity (\texttt{fog}) & $8$ \tabularnewline
\hline 
\paramPrecipitation (\texttt{rain}) & $51$ \tabularnewline
\hline 
\paramSunAzimuthAngle & $90$ \tabularnewline
\hline 
\paramSunAltitudeAngle (\texttt{sun}) & $54$ \tabularnewline
\hline 
\paramWindIntensity (\texttt{wind}) & $43$ \tabularnewline
\hline 
\paramPrecipitationDeposits (\texttt{deposit}) & $76$ \tabularnewline
\hline 
\end{tabular}\hfill{}~
\par\end{center}%
\end{minipage}
\par\end{centering}
}
\par\end{centering}
\begin{centering}
\subfloat[Histograms for the 6 marginals of the predicted weather distribution for $\anImage$.\label{fig:task-I-histograms}]{
\begin{centering}
\noindent\begin{minipage}[t]{1\textwidth}%
\begin{center}
~\hfill{}\includegraphics[width=0.3\textwidth]{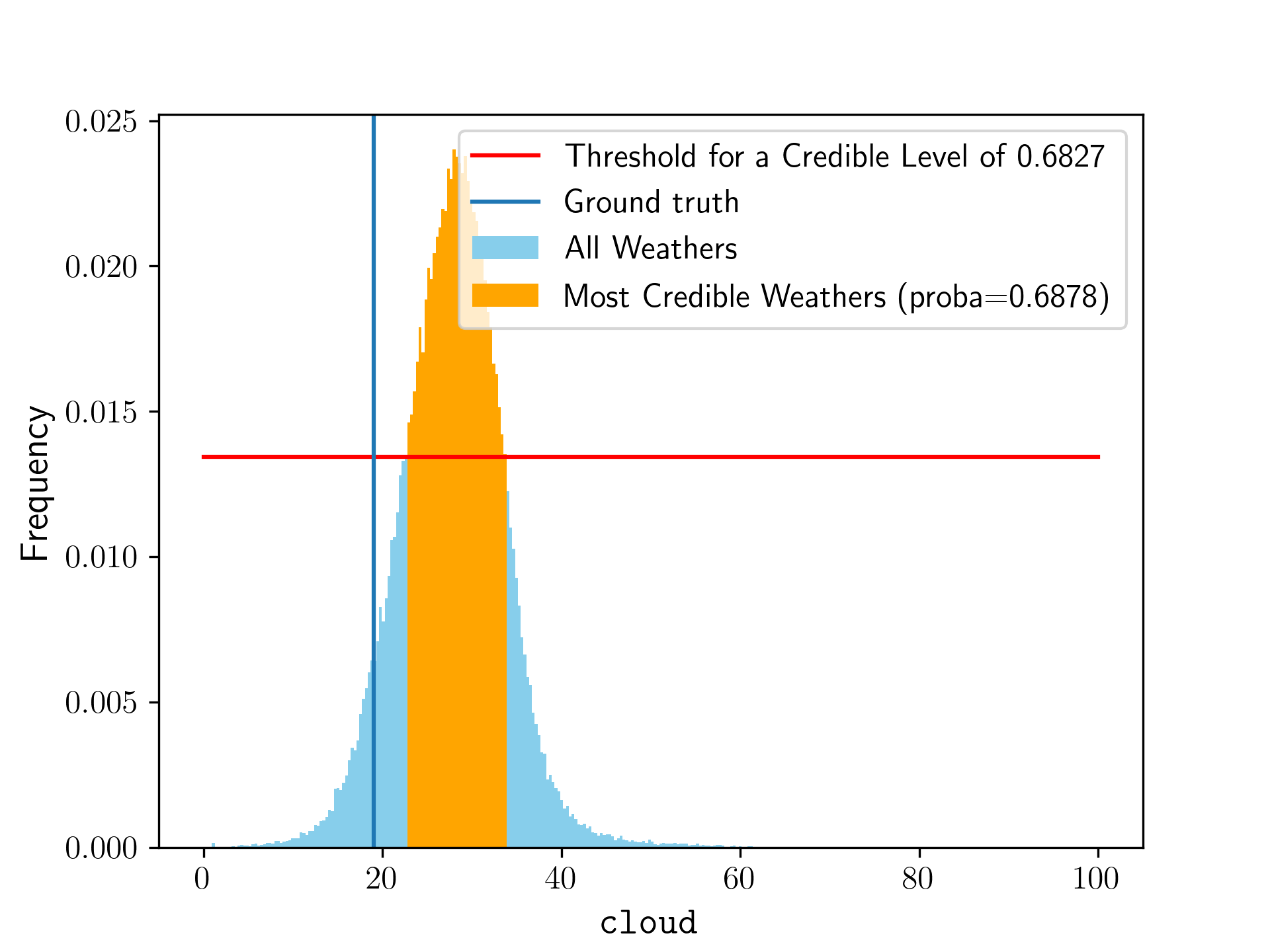}\hfill{}\hfill{}\includegraphics[width=0.3\textwidth]{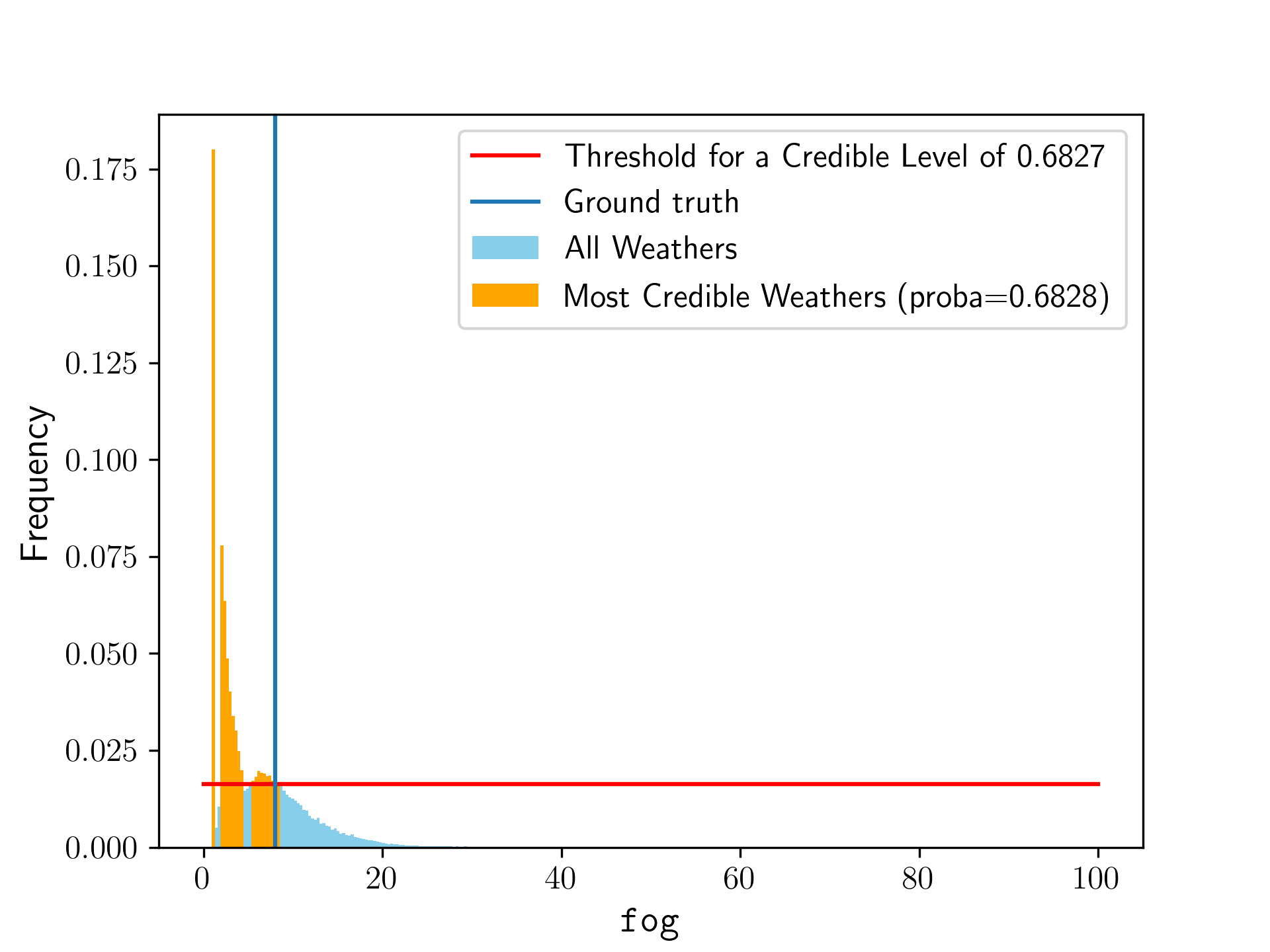}\hfill{}\hfill{}\includegraphics[width=0.3\textwidth]{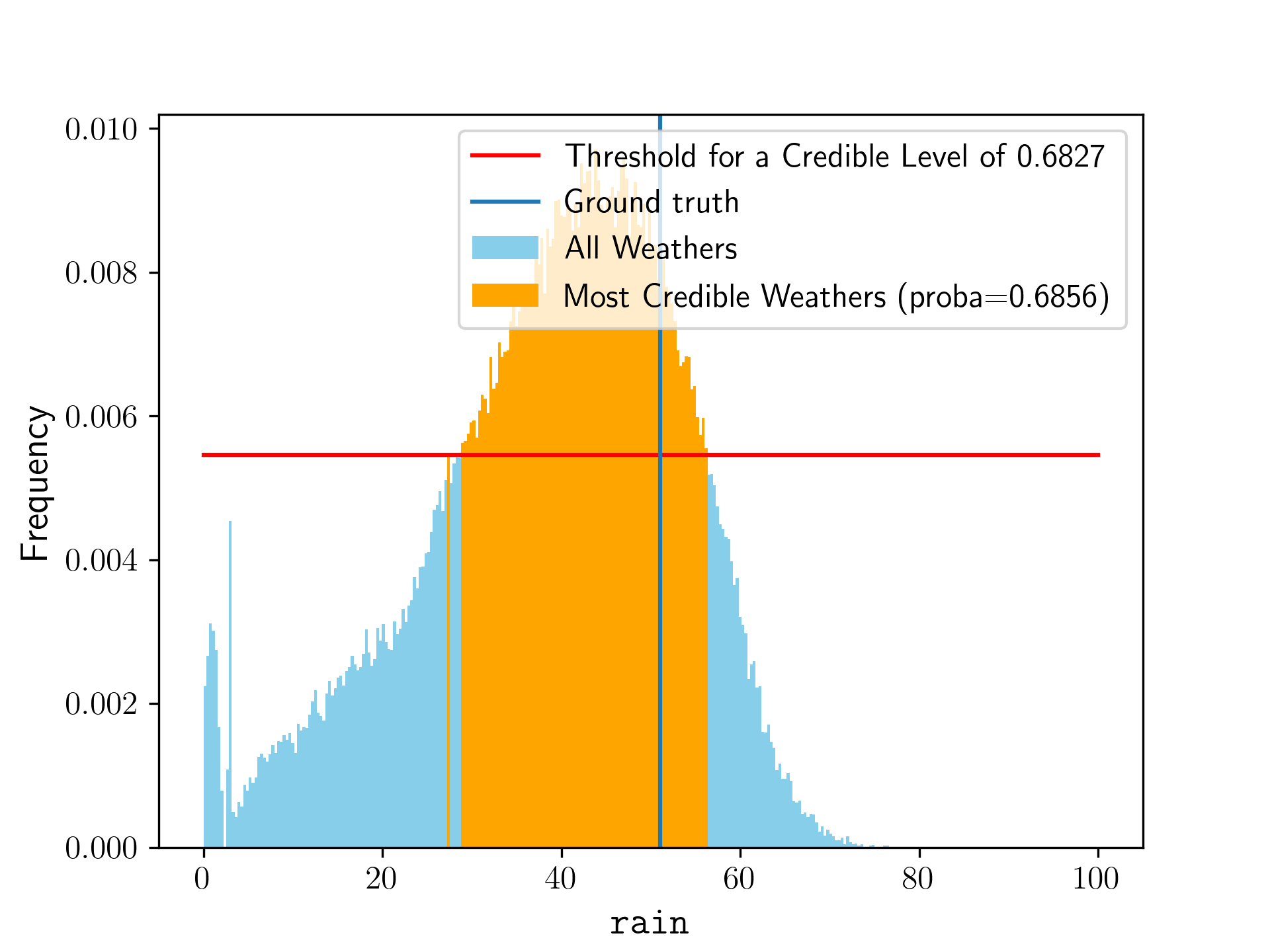}\hfill{}~
\par\end{center}
\vspace{-8mm}
\begin{center}
~\hfill{}\includegraphics[width=0.3\textwidth]{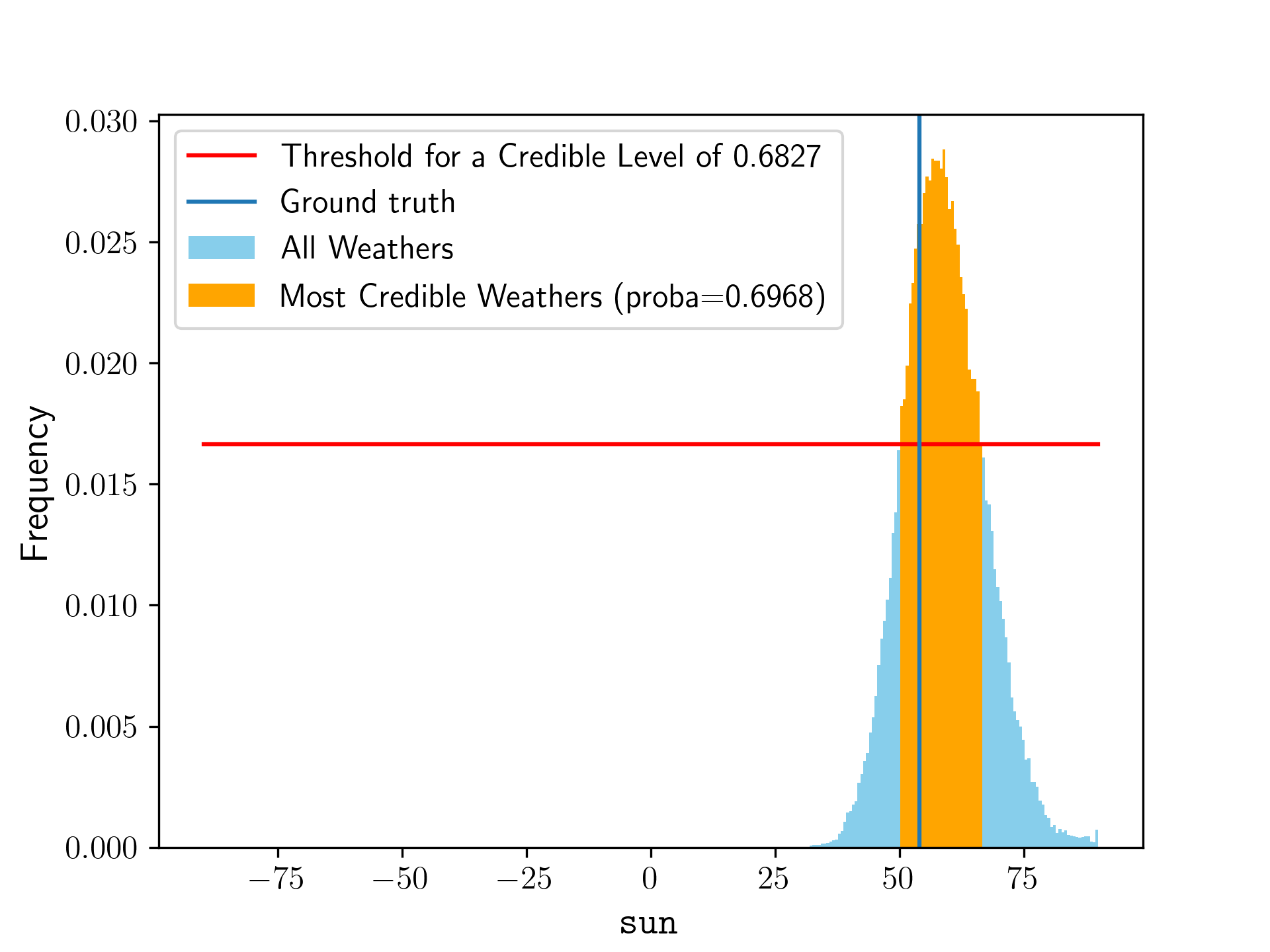}\hfill{}\hfill{}\includegraphics[width=0.3\textwidth]{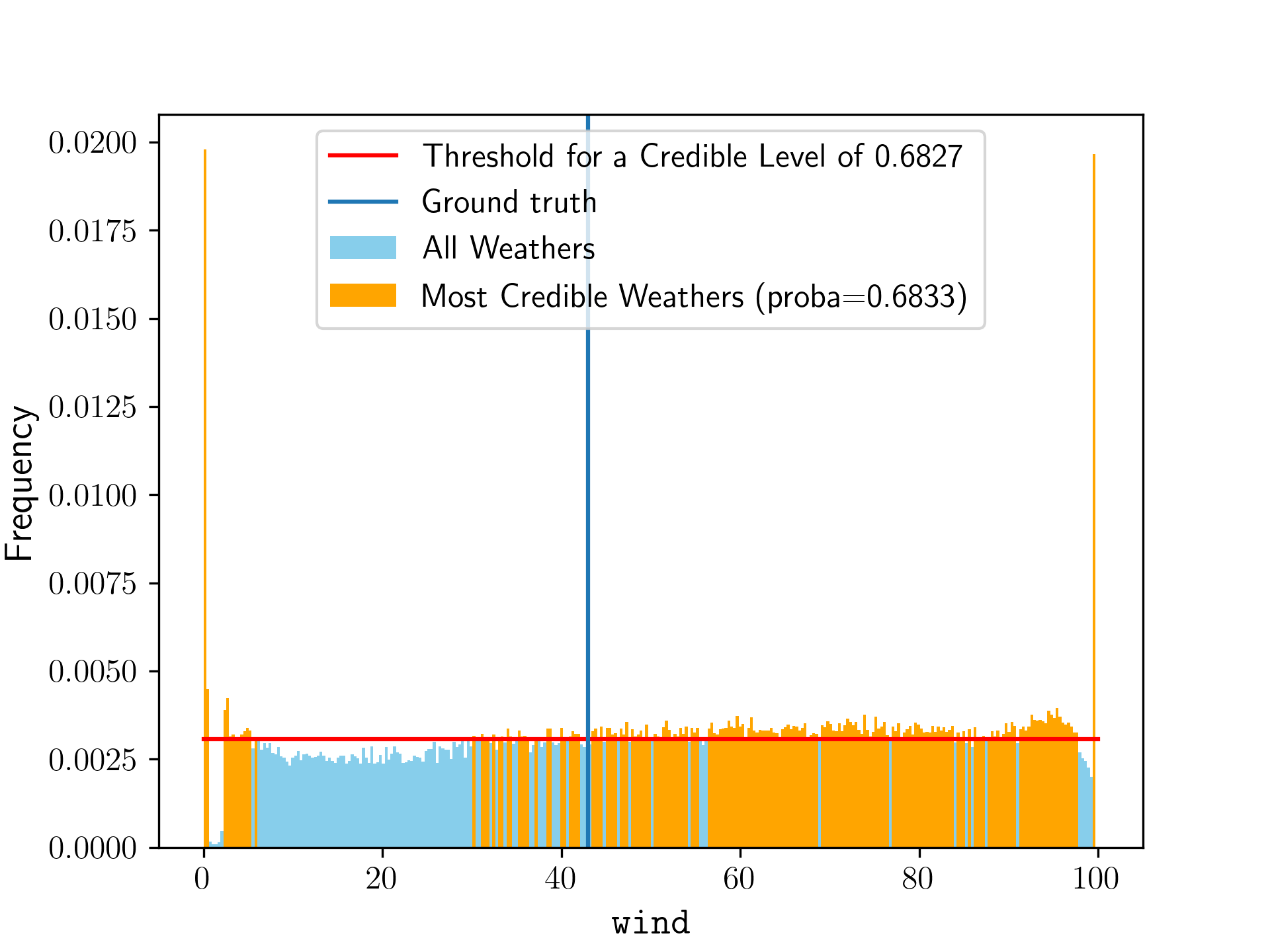}\hfill{}\hfill{}\includegraphics[width=0.3\textwidth]{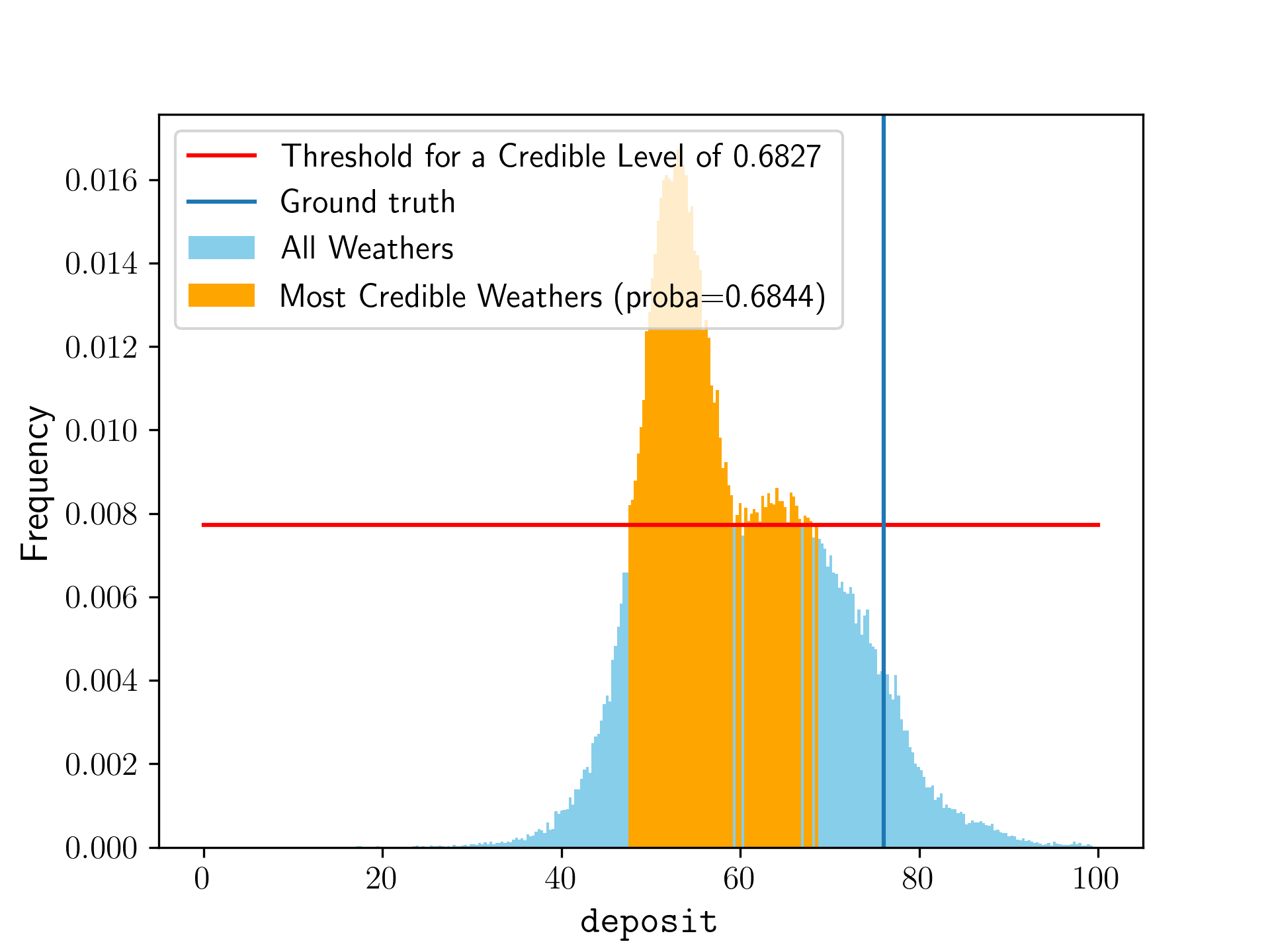}\hfill{}~
\par\end{center}%
\end{minipage}
\par\end{centering}
}
\par\end{centering}
\begin{centering}
\subfloat[Corner plot, with the ground-truth weather pinned (left: complete, right: zoomed).\label{fig:task-I-corner-plots}]{
\begin{centering}
\noindent\begin{minipage}[t]{1\textwidth}%
\begin{center}
~\hfill{}\hfill{}\includegraphics[width=0.30\textwidth]{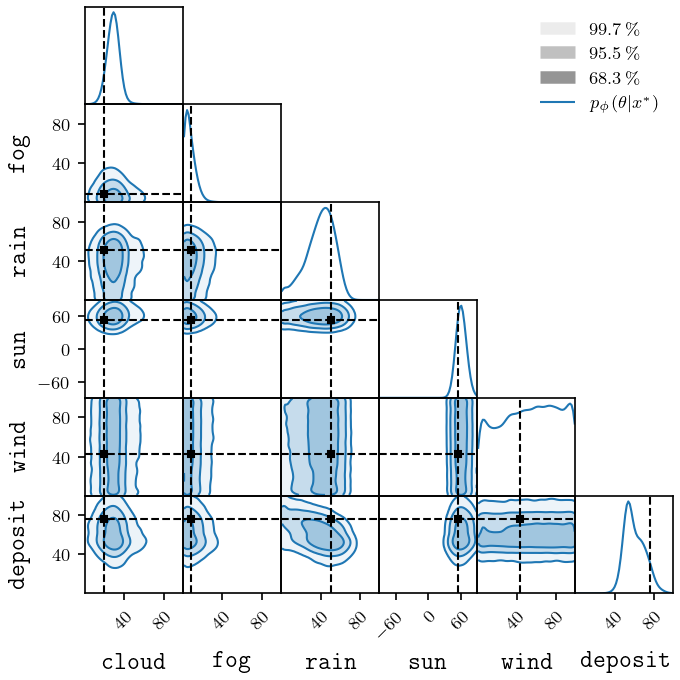}\hfill{}\includegraphics[width=0.30\textwidth]{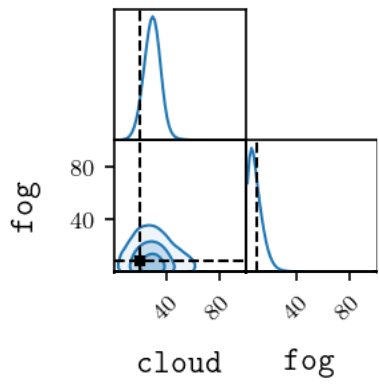}\hfill{}\hfill{}~
\par\end{center}%
\end{minipage}
\par\end{centering}
}
\par\end{centering}
\begin{centering}
\subfloat[Posterior predictive check.\label{fig:task-I-ppc}]{
\begin{centering}
\noindent\begin{minipage}[t]{1\textwidth}%
\begin{center}
\includegraphics[width=1\textwidth]{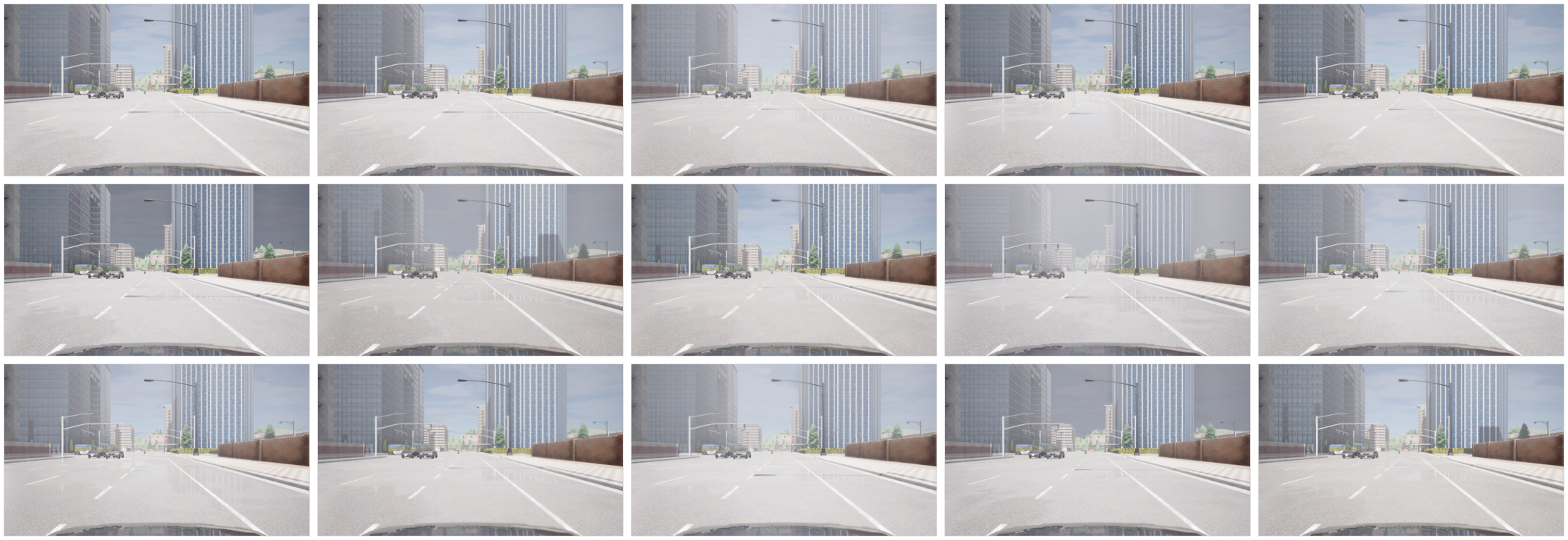}
\par\end{center}%
\end{minipage}
\par\end{centering}
}
\par\end{centering}
\caption{Task I: results obtained, with the model learned with the \resnet backbone on $500$k learning samples, for an arbitrarily chosen input image in the \TS.}

\end{figure}

\paragraph{1st analysis: histograms (Fig.~\ref{fig:task-I-histograms}).}
Due to their physical meaning, the weather parameters are easily interpretable. This paves the way to a first, subjective, evaluation. We draw samples at random out of the predicted weather distributions $\hat{\probabilityMeasure}_{\domainRef}(\rvWeather\vert \rvImage=\anImage)$, for some images $\anImage$ arbitrarily chosen in \TS and conduct a visual inspection of the histograms for the $6$ marginals. The most credible weather parameters (\aka highest density credibility sets, highest density regions, plausible sets, \etc) are highlighted for a credible level $\credibleLevel=68.27\%$. For any input image $\anImage$, 
these weathers are those such that the predicted \pdf is above some threshold $t(\anImage)$ and the predicted probability of the set is $\credibleLevel$~\cite{Hyndman1996Computing}.

\paragraph{2nd analysis: corner plots (Fig.~\ref{fig:task-I-corner-plots}).}
A corner plot is a triangular array of plots. Those on the diagonal show the $n$ marginals of a $n$-D distribution. Those below the diagonal depict the highest density credibility regions delimited at some arbitrarily chosen credibility levels (in this paper: 68.27\%, 95.45\%, and 99.73\%), for each pair of parameters. We observe on the corner plots wide distributions, meaning that there is a large uncertainty for the weather parameters given an image. However, when analyzing the results, this uncertainty is explainable and the following analysis shows that this uncertainty is not excessive, \ie, our models are not underconfident.

\paragraph{3rd analysis: posterior predictive checks (Fig.~\ref{fig:task-I-ppc}).}
Once a weather distribution is predicted for an observation (input image), it is possible to (1)~draw weather parameters vectors at random from it and then (2)~to inject these vectors in \carla, keeping all other parameters unchanged and immobilizing the vehicles and pedestrians, to obtain new images in order to finally (3)~compare those with the initial input image. One cannot expect to obtain identical images as, in \carla, the traffic lights continue to run, the rain continues to fall, the plants move with the wind, and pedestrians' poses are not perfectly frozen. Putting this aside, we observe that most retrieved images are very similar to the input image. 
We conclude that our models are not underconfident. 

\begin{figure}[t!]
    \centering
    ~\hfill
    \includegraphics[height=0.3\textwidth]{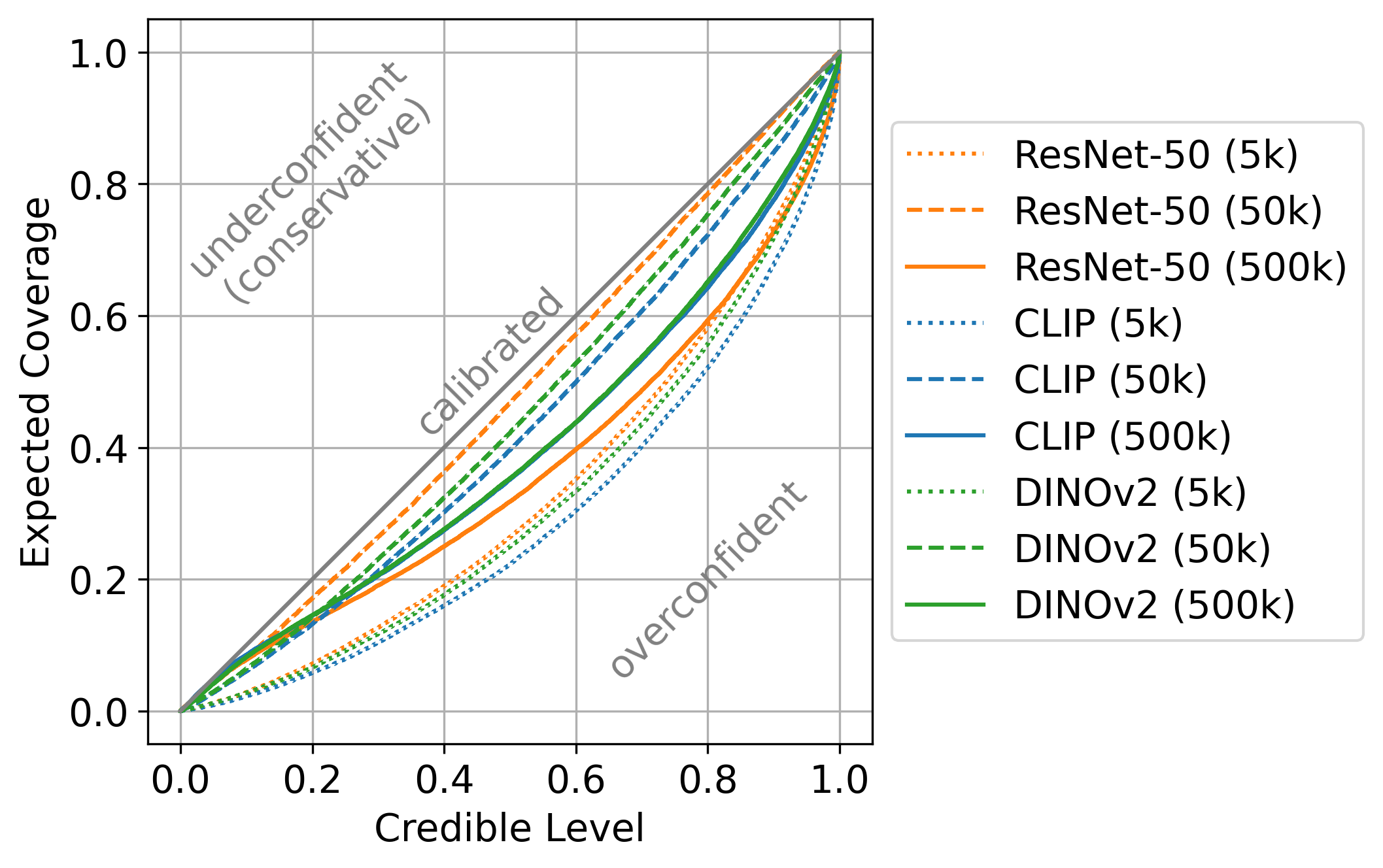}
    \hfill\hfill
    \includegraphics[height=0.3\textwidth]{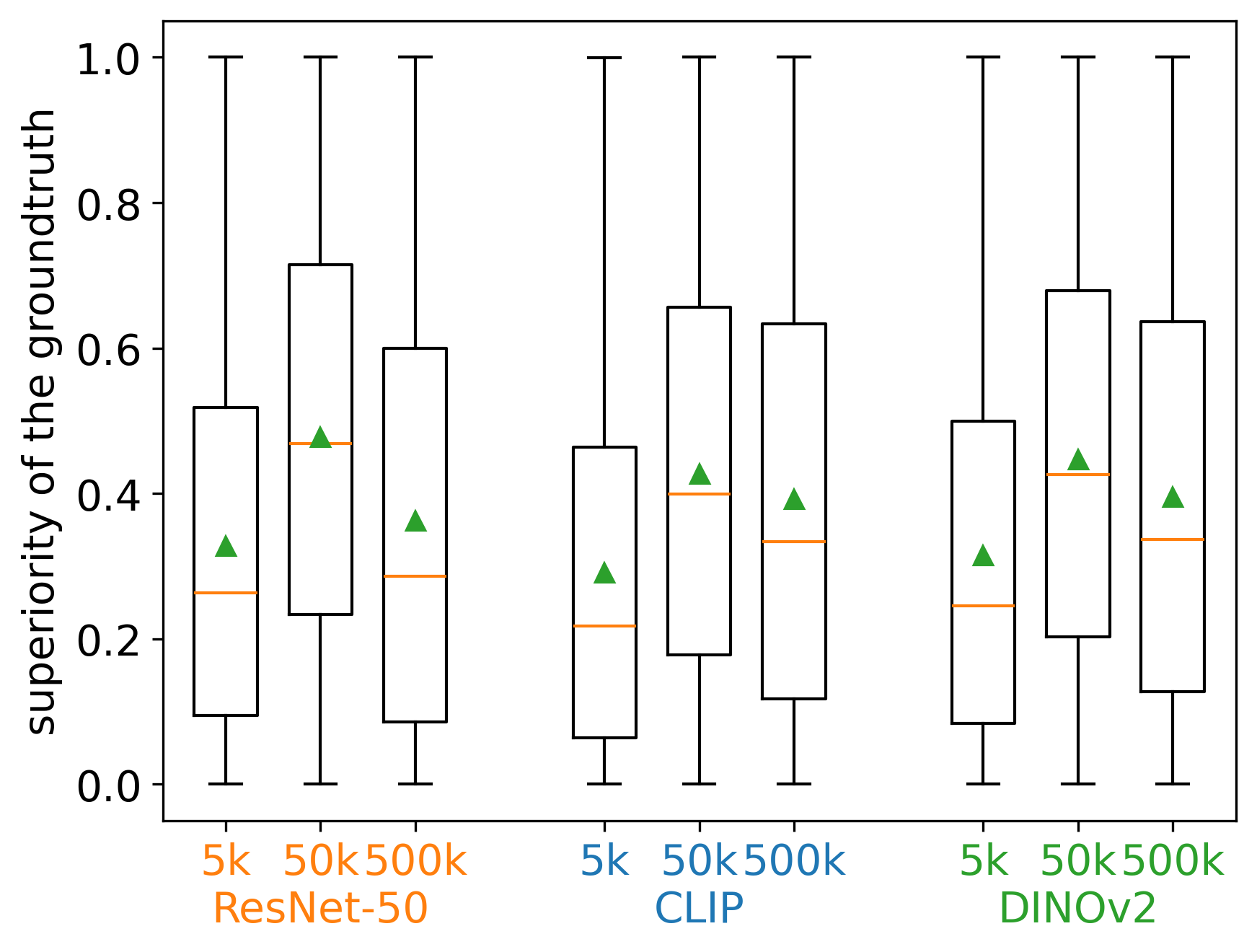}
    \hfill~
    \caption{Task I: comparison of our 9 models on the overall \TS. Left: coverage plot. Right: box-and-whisker plots for the $\superiorityGT$ statistic. Both analyses agree that the model learned with the \resnet backbone on $50$k learning samples is preferable.}
    \label{fig:task-I-coverage-plot}\label{fig:task-I-superiority-gt}
\end{figure}

\paragraph{4th analysis: coverage plots (Fig.~\ref{fig:task-I-coverage-plot}).}
Coverage plots show how the expected coverage varies with the credible level~\cite{hermans2021trust}. By definition, the expected coverage, at the credible level $\credibleLevel\in [0,1]$, is the probability that the ground-truth weather belongs to the most credible weathers at level $\credibleLevel$. A method estimating the distribution of weathers based on an image is said underconfident (\ie, conservative), calibrated, or overconfident at level $\credibleLevel$ when the expected coverage at level $\credibleLevel$ is, respectively, $> \credibleLevel$, $= \credibleLevel$, or $< \credibleLevel$.  We observe that our models are all overconfident. The best calibrated model is the one that we obtained with $50$k learning samples and \resnet as backbone.

\paragraph{5th analysis: superiority of the ground truth (Fig.~\ref{fig:task-I-superiority-gt}).} We also determine to what extent the ground truth is more credible than the other weathers. This is the proportion $\superiorityGT$ of weathers that can be drawn at random from the normalizing flow and that are predicted as having a \pdf value (\ie, likelihood) lower or equal than the one from the ground truth. The higher $\superiorityGT$ is, the better it is, but approaching $1.0$ is notably very challenging. We estimated $\superiorityGT$ for each image of the \TS and made box-and-whisker plots for the $9$ models. This analysis is complementary to the coverage plot in the sense that a model can be perfectly calibrated while still presenting room for improvement. That being said, we are in the particular case in which this analysis leads to the same conclusion as the coverage plot: the model based on the backbone \resnet and learned from $50$k samples is preferable to the others.

\subsection{Task II: Absolute Domain Characterization}
\label{sec:task-2}
\begin{figure}[b!]
  \centering
    \includegraphics[width=\linewidth]{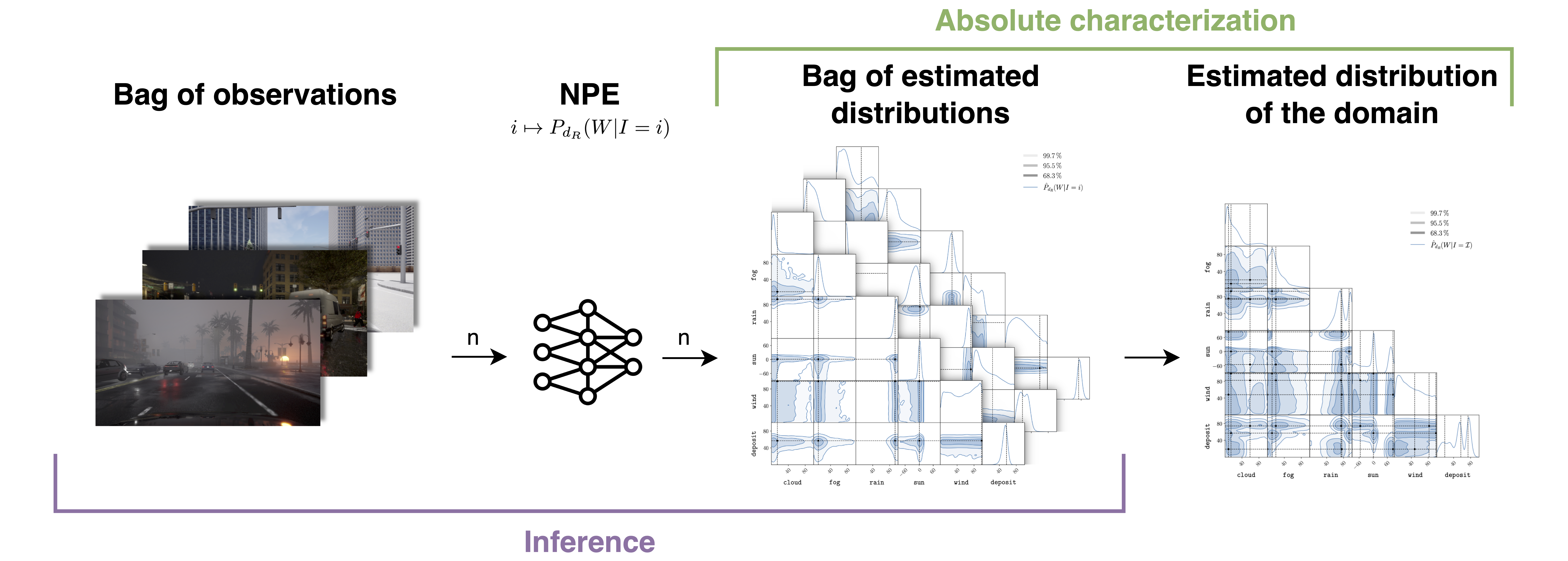}
  
  \caption{
  Task II.
  The aim of the second task is to obtain an absolute characterization of a domain of interest. We process with the \npe each observation of a bag of observations of the domain of interest to obtain the bag of estimated distributions corresponding to the observations, then we obtain the distribution of weather parameters for the domain of interest by averaging the individual distributions of the bag.
  }\label{fig:task-II} 
\end{figure}

The second task (see Fig.~\ref{fig:task-II}) consists in characterizing a domain in an easy-to-interpret way. For this, we opt for a distribution of physical parameters estimated based on a bag (\aka multiset) of observations.

\subsubsection{Case Study.}
We characterize a domain $\domain\in\allDomains$ by the estimated distribution $\hat{\probabilityMeasure}_\domain (\rvWeather)$ of weather conditions based on a real-valued bag $\aBag$ (multiset) of arbitrarily weighted images acquired by cameras placed in front of vehicles. In the following, we denote the weight (multiplicity) of the image $\anImage\in\aBag$ by $\weight(\anImage)$. The images can either originate from a unique vehicle or from several, in case of vehicle-to-vehicle (\vtov) communications. We want to establish 
$
    \aBag \mapsto \hat{\probabilityMeasure}_\domain(\rvWeather\vert \rvBag=\aBag)
$.

\subsubsection{Data.}
We consider a bag $\aBag$ of $1{,}000$ equally weighted images generated 
in maps already used in Task~I. The  weather parameters have been drawn at random from an arbitrary distribution $\probabilityMeasure_\domain (\rvWeather)$ for which we fixed all parameters but two, \paramFogDensity and \paramPrecipitation, that follow a mixture of Gaussians.

\subsubsection{Method.}
We use the \npe model developed for task I and build our solution for task II on top of it.  We implement the following estimator for $\probabilityMeasure_\domain(\rvWeather)$: 
$
    \hat{\probabilityMeasure_\domain} (\rvWeather) = \sum_{\anImage \in \aBag} \weight(\anImage) \hat{\probabilityMeasure}_{\domainRef}(\rvWeather\vert \rvImage=\anImage)
$. 
The motivation for this estimator is threefold. (1)~It is straightforward to evaluate the probability density function of $\hat{\probabilityMeasure_\domain}(\rvWeather)$ and to draw weather conditions $\aWeather\in\allWeathers$ at random, following $\hat{\probabilityMeasure_\domain}(\rvWeather)$, as we can do it with the normalizing flow $\probabilityMeasure_{\domainRef}(\rvWeather\vert \rvImage=\anImage)$. 
This will be valuable in our third task. (2)~This estimator is useful for linear temporal filtering, \eg, when one wants to weight more the recently acquired images than the old ones. (3)~Finally, this estimator is fully justifiable under the assumptions that $\probabilityMeasure_\domain (\rvWeather\vert\rvImage) = \probabilityMeasure_{\domainRef} (\rvWeather\vert\rvImage)$ and $\weight(\anImage)=\probabilityMeasure_\domain (\rvImage=\anImage)$ as, in this case, $\probabilityMeasure_\domain (\rvWeather)=\int_{\anImage} \weight(\anImage) \probabilityMeasure_{\domainRef} (\rvWeather\vert\rvImage=\anImage) d\anImage$.

\subsubsection{Evaluation and Results.}
\begin{figure}[t!]
    \centering
    \includegraphics[height=0.28\textwidth]{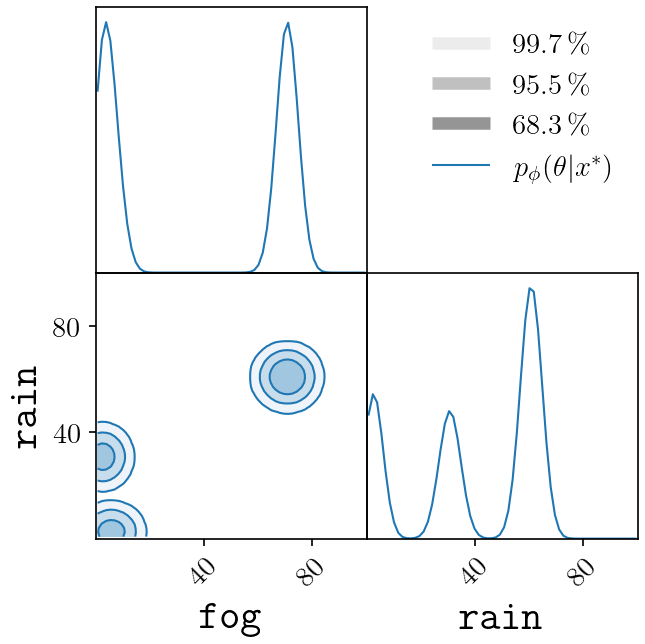}
    \hfill
    \includegraphics[viewport=0bp 0bp 1322bp 1008bp,clip,height=0.28\textwidth]{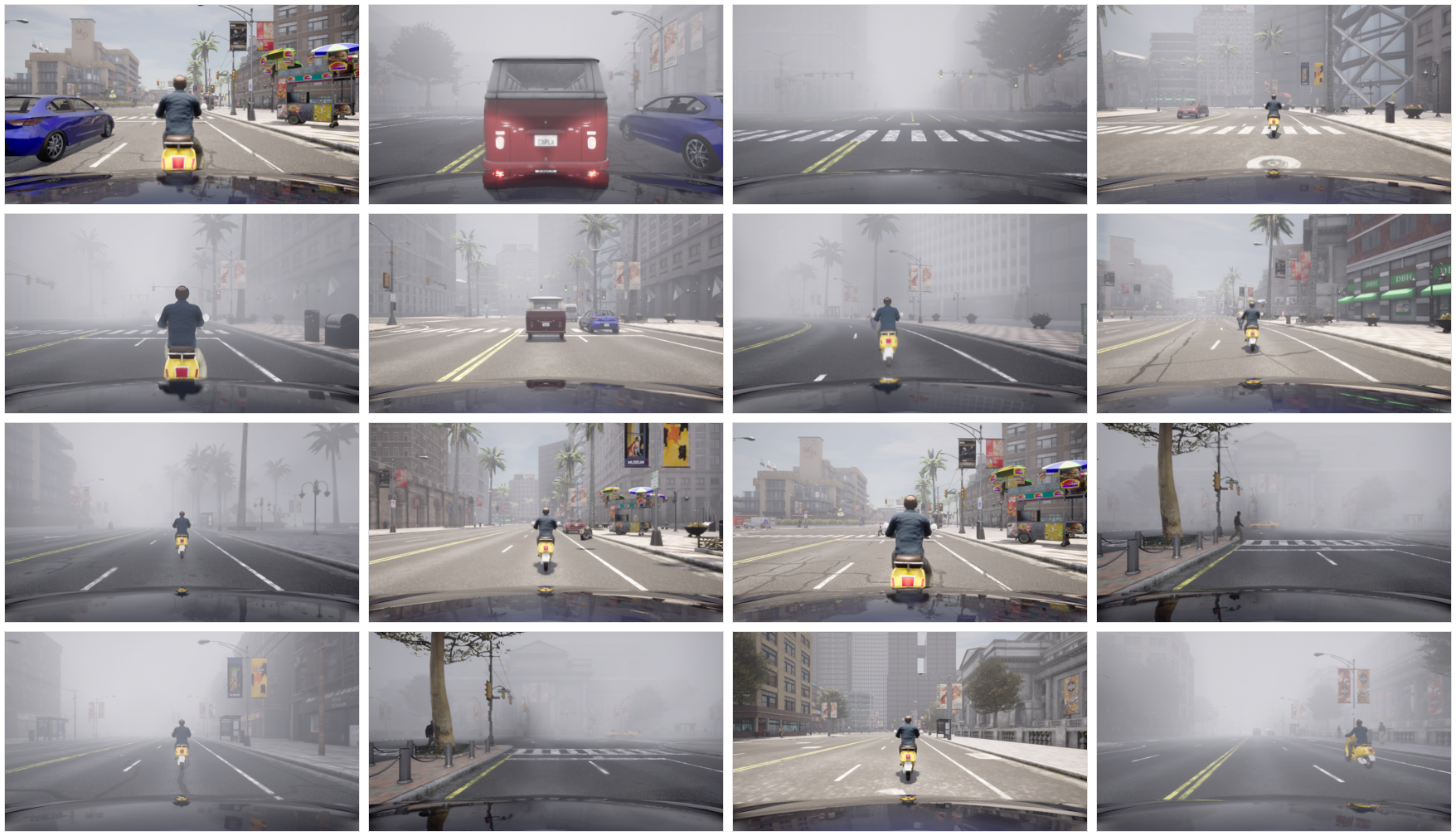}
    \hfill
    \includegraphics[height=0.28\textwidth]{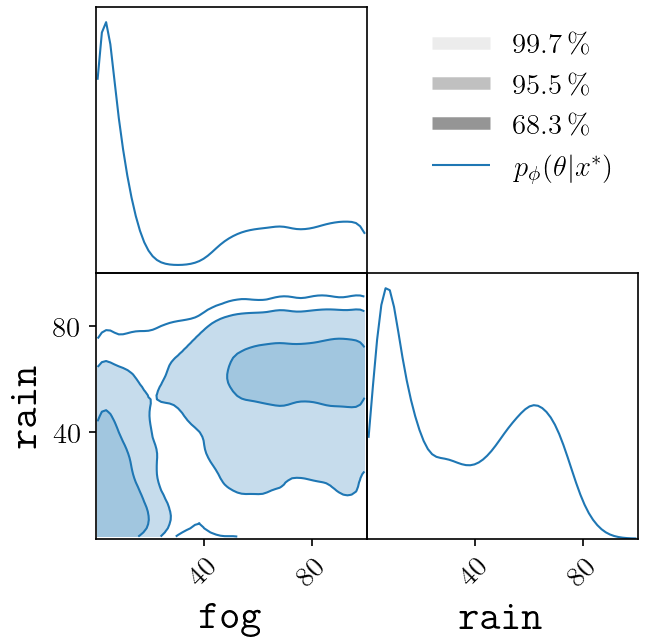}
    \caption{Task II: example of result. Left: corner plot showing an arbitrarily chosen distribution of weather conditions, $\probabilityMeasure_\domain(\rvWeather)$ (ground truth). Center: a bag $\aBag$ of images generated using \carla with weather conditions drawn from $\probabilityMeasure_\domain(\rvWeather)$. Right: corner plot showing the estimated likelihood of the weather conditions $\hat{\probabilityMeasure}_\domain(\rvWeather)$ based on $\aBag$.}\label{fig:task-II-result-rain-fog}
\end{figure}

Our result is shown in Fig.~\ref{fig:task-II-result-rain-fog}. We observe that the three modes of the ground-truth weather distribution $\probabilityMeasure_\domain(\rvWeather)$ are within the highest density regions of the prediction $\hat{\probabilityMeasure}_\domain(\rvWeather)$. We made the same observation for many other ground-truth distributions (results omitted due to the limited space). We stress the fact that $\hat{\probabilityMeasure}_\domain(\rvWeather)$ differs significantly from $\probabilityMeasure_\domain(\rvWeather)$ can be explained by the inherent loss of information resulting from the use of color cameras, as already discussed in the introduction (\cf Fig.~\ref{fig:observation-1-ambiguous-cases}).

\subsection{Task III: Relative Domain Characterization}
\label{sec:task-3}
\begin{figure}[t!]
  \centering
    \includegraphics[width=\linewidth]{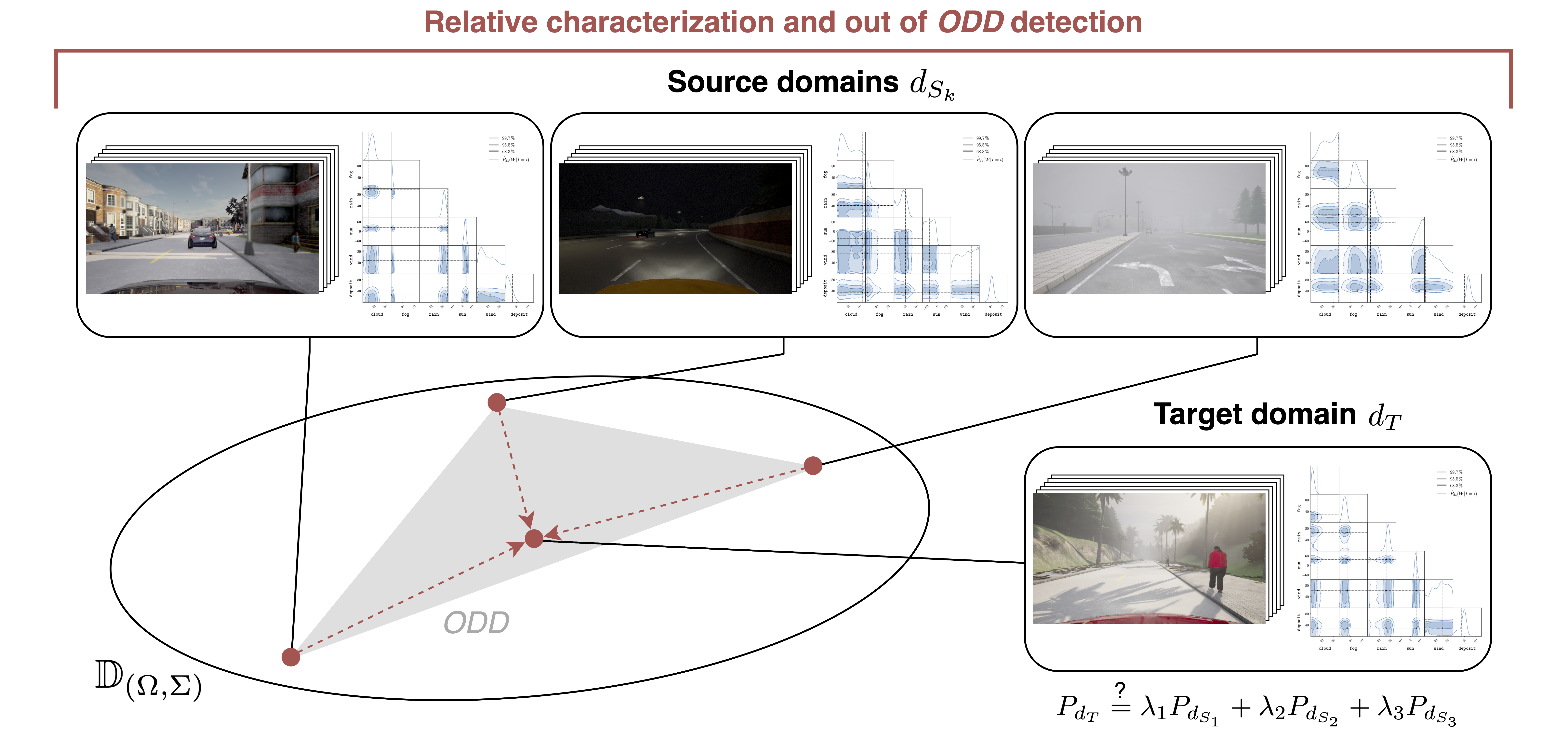}
  
  \caption{Task III. The aim of the third task is twofold: (1)~having a relative characterization of a target domain $\domainTarget$ based on source domains $\domainSource[k]$ and (2)~detecting when this target domain $\domainTarget$ is out of the \odd.}\label{fig:task-III}
\end{figure}

The third task (see Fig.~\ref{fig:task-III}) consists in characterizing any target domain $\domainTarget\in\allDomains$, relatively, \wrt to some arbitrarily chosen source domains $\domainSource[1]$, $\domainSource[2]$, \ldots $\domainSource[\numSources]\in\allDomains$. Our motivation for this task originates from the importance of knowing if a system implementing a given many-to-one domain adaptation method can operate safely in the target domain $\domainTarget$ when it is known to operate safely in the source domains $\domainSource[1],\domainSource[2],\ldots\domainSource[\numSources]$.

\subsubsection{Case Study.}
We discuss the case of \emph{Mixture Domain Adaptation} (\mda) in which the system adapts to any target domain for which 
$\probabilityMeasure_{\domainTarget} = \sum_{k=1}^{\numSources} \lambda_k \probabilityMeasure_{\domainSource[k]}$, with $\sum_{k=1}^{\numSources} \lambda_k = 1$ and $\lambda_k \ge 0 \, \forall k$ (this is the \emph{mixture assumption}). In this framework, the \odd is the convex hull of $\{\probabilityMeasure_{\domainSource[1]},\probabilityMeasure_{\domainSource[2]},\ldots\probabilityMeasure_{\domainSource[\numSources]}\}$. An autonomous car implementing \mda is expected to drive safely when $\probabilityMeasure_{\domainTarget}\in\odd$. So, our goal in this third task is to determine if the mixture assumption holds and if so, what the values of the mixture weights $\lambda_k$ are.

\mda supports many potential applications in the high-level tasks of the \emph{Sense} pillar of the \emph{Sense-Plan-Act} model~\cite{Shoker2024Savvy}. While Mansour studied it generically~\cite{Mansour2008Domain}, the application to the two-class classification task has been studied in~\cite{Pierard2014OnTheFly}, and the application to the semantic segmentation task of images acquired by vehicle-mounted cameras has been studied in~\cite{Pierard2023Mixture}. These latter two works put an important emphasis on on-the-fly applicability and provide mathematically proven exact solutions. However, a critical limitation of these works is the need for the mixture weights to be known at adaptation time. Here, we remove this limitation by introducing a method that determines these weights automatically.

\subsubsection{Data.} We consider $4$ subsets of the \TS that we created for Task~I: $\aBag_0$, $\aBag_1$, $\aBag_2$, and $\aBag_3$, containing $176$, $27$, $21$, and $26$ images, respectively. In the bag $\aBag_k$, the distribution of the ground-truth weather parameters follows a uniform distribution on $\allWeathers_k\subsetneq\allWeathers$. These sets are such that $\allWeathers_0\cap(\allWeathers_1\cup\allWeathers_2\cup\allWeathers_3)=\emptyset$.

\subsubsection{Method.}
\newcommand{\meanSquareGap}{\delta(\hat{\lambda}_1,\ldots\hat{\lambda}_{\numSources})}
We characterize the source and target domains with the method presented for Task~II, using the same weather distribution predictive model for all domains. We define the mean squared gap between the target domain and the mixture of the source domains as:

\begin{eqnarray}
    \meanSquareGap
    &=
    \int_{\allWeathers}
    \left[
    \hat{\probabilityMeasure}_{\domainTarget}(\rvWeather=\aWeather)
    -
    \sum_{k=1}^{\numSources} \hat{\lambda}_k \hat{\probabilityMeasure}_{\domainSource[k]}(\rvWeather=\aWeather)
    \right]^2
    \hat{\probabilityMeasure}_{\domainTarget}(\rvWeather=\aWeather) d\aWeather
    \nonumber 
    \\
    &\simeq
    \frac{1}{\numWeathers}
    \sum_{i}^{\numWeathers}
    \left[
    \hat{\probabilityMeasure}_{\domainTarget}(\rvWeather=\aWeather_i)
    -
    \sum_{k=1}^{\numSources} \hat{\lambda}_k \hat{\probabilityMeasure}_{\domainSource[k]}(\rvWeather=\aWeather_i)
    \right]^2
\end{eqnarray}
with
$\{\aWeather_i\}_{i=1}^{\numWeathers}\sim\hat{\probabilityMeasure}_{\domainTarget}(\rvWeather)$. We aim at finding the values of $\hat{\lambda}_1,\ldots\hat{\lambda}_{\numSources}$ that minimize $\delta(\hat{\lambda}_1,\ldots\hat{\lambda}_{\numSources})$. This is a constrained least squares problem in which the constraints are $\sum_{k=1}^{\numSources} \hat{\lambda}_k = 1$ and $\hat{\lambda}_k \ge 0 \, \forall k$.
In order to use the \cvxpy library~\cite{Diamond2016CVXPY, Agrawal2018Rewriting} to solve our problem, we converted our original problem into a convex quadratic programming problem, with the same constraints.

\subsubsection{Evaluation and Results.}
The goal of this experiment is twofold.
\begin{enumerate}
    \item To show that it is possible to recover the mixture weights needed for the \mda technique presented in~\cite{Pierard2023Mixture}. If the probability measure in the target domain is a mixture of the probability measures in the source domains, then we expect our characterization of the target domain to be a mixture of our characterizations for the source domains, with the same weights, as
    $
    \probabilityMeasure_{\domainTarget}=\sum_{k=1}^{\numSources} \lambda_k \probabilityMeasure_{\domainSource[k]}
    \,\Rightarrow\,
    \probabilityMeasure_{\domainTarget}(\rvWeather)=\sum_{k=1}^{\numSources} \lambda_k \probabilityMeasure_{\domainSource[k]}(\rvWeather)
    $. 
    \item To show that it is possible to detect when the target domain is not a mixture of the source domains, which means that the target domain is out of the \odd for the \mda technique presented in~\cite{Pierard2023Mixture}. If our characterization of the target domain is not a mixture of our characterizations for the source domains, then we expect that the probability measure in the target domain is not a mixture of the probability measures in the source domains, as we have
    $
    \probabilityMeasure_{\domainTarget}(\rvWeather)\ne\sum_{k=1}^{\numSources} \lambda_k \probabilityMeasure_{\domainSource[k]}(\rvWeather)
    \,\Rightarrow\,\probabilityMeasure_{\domainTarget}\ne\sum_{k=1}^{\numSources} \lambda_k \probabilityMeasure_{\domainSource[k]}
    $.
\end{enumerate}

\newcommand{\noiseLevel}{\eta}
To achieve this goal, we perform experiments in which we characterize target domains relatively to $\numSources=3$ source domains. However, to study the robustness of our method, we let the probability measures in the target domains $\domainTarget$ be mixtures not only of the probability measures in the source domains, but also in some domains that are out of the \odd. We use the bags $\aBag_1$, $\aBag_2$, and $\aBag_3$ for the source domains $\domainSource[1]$, $\domainSource[2]$, and $\domainSource[3]$, respectively. We also use the bag $\aBag_0$ for domains $\domainOutODD$ that are, by construction, guaranteed to be out of the \odd. Putting this into equations, we have $\probabilityMeasure_{\domainTarget}
=(1-\noiseLevel) \probabilityMeasure_{\domainInODD}+\noiseLevel \probabilityMeasure_{\domainOutODD}$ with $\probabilityMeasure_{\domainInODD} 
= \lambda_1 \probabilityMeasure_{\domainSource[1]}
+ \lambda_2 \probabilityMeasure_{\domainSource[2]}
+ \lambda_3 \probabilityMeasure_{\domainSource[3]}$. The quantity $\noiseLevel\in[0,1]$ is interpreted as a proportion of noise, which is the keystone to assess the robustness of our method.

In each experiment, the images in $\aBag_1$, $\aBag_2$, and $\aBag_3$ are equally weighted, whereas those in $\aBag_0$ are randomly weighted following a uniform distribution. 
The method introduced in Task~II for the absolute characterization of domains is applied for $\domainSource[1]$, $\domainSource[2]$, $\domainSource[3]$, and $\domainTarget$. We arbitrarily chose $(\lambda_1,\lambda_2,\lambda_3)=(0.2, 0.3, 0.5)$. The mixture weights are optimized on $\numWeathers=16$ points.

\newcommand{\euclideanError}{d_E}
The overall experiment consists in (1)~choosing a target domain at random, as described here-above, (2)~computing its absolute characterization with the method of Task~II, (3)~executing the weight estimation algorithm, and (4)~reporting both the mean square gap $\meanSquareGap$ and the Euclidean distance $\euclideanError$ between $(\hat{\lambda}_1,\hat{\lambda}_2,\hat{\lambda}_3)$ and $(\lambda_1,\lambda_2,\lambda_3)$. This experiment has been performed $330$ times ($30$ times for $11$ values of $\noiseLevel$). The results are shown in Fig.~\ref{fig:task-III-results}. The mean square gap achievable by chance is shown (baseline). Note that the target domain belongs to the \odd only when $\noiseLevel=0$. We observe that $\euclideanError$ is negligible when $\domainTarget\in\odd$ and that the mean square gap is effective to detect when $\domainTarget\notin\odd$.

\begin{figure}[t!]
    \centering
    ~\hfill
    \includegraphics[width=0.45\textwidth]{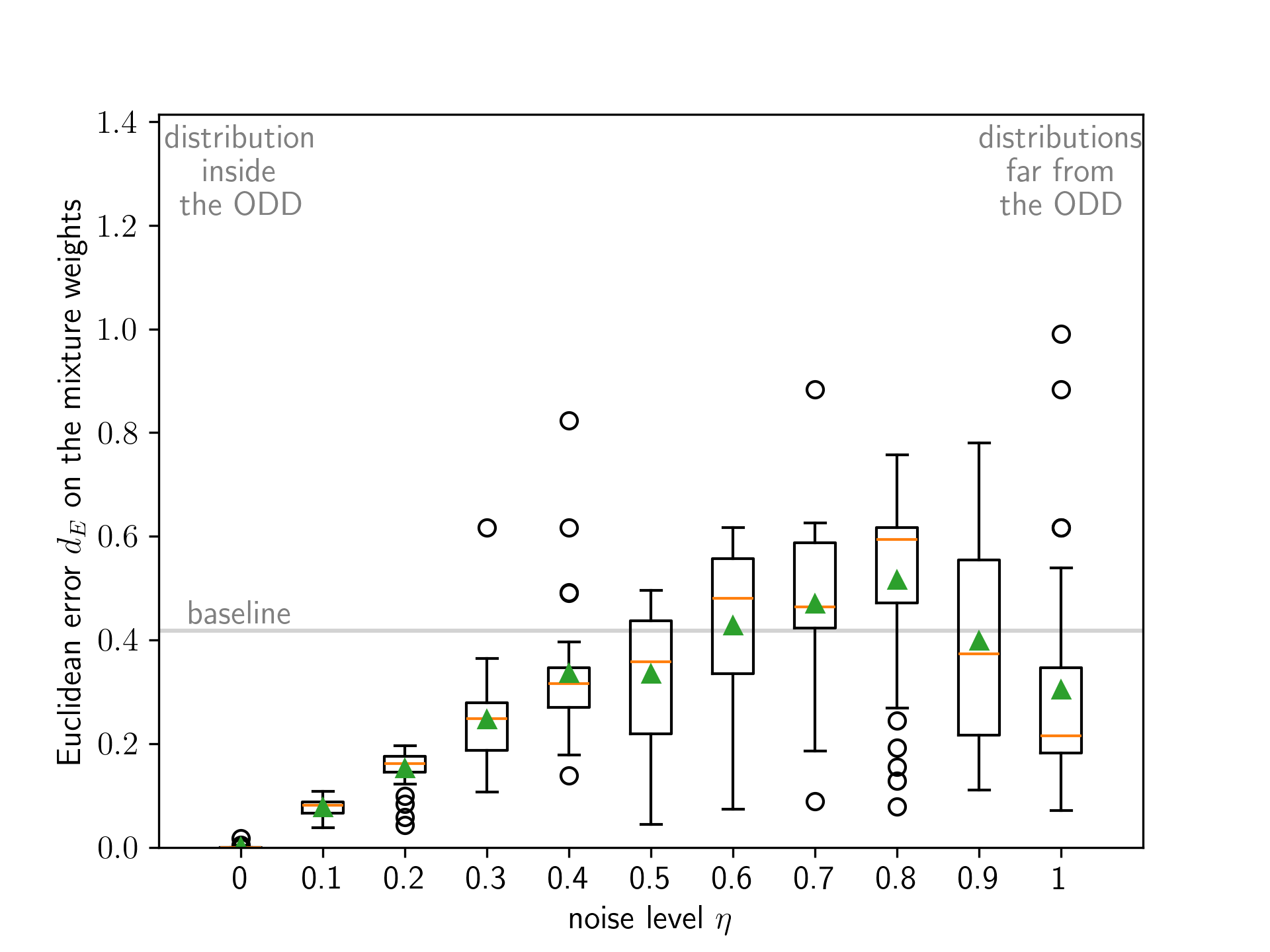}
    \hfill\hfill
    \includegraphics[width=0.45\textwidth]{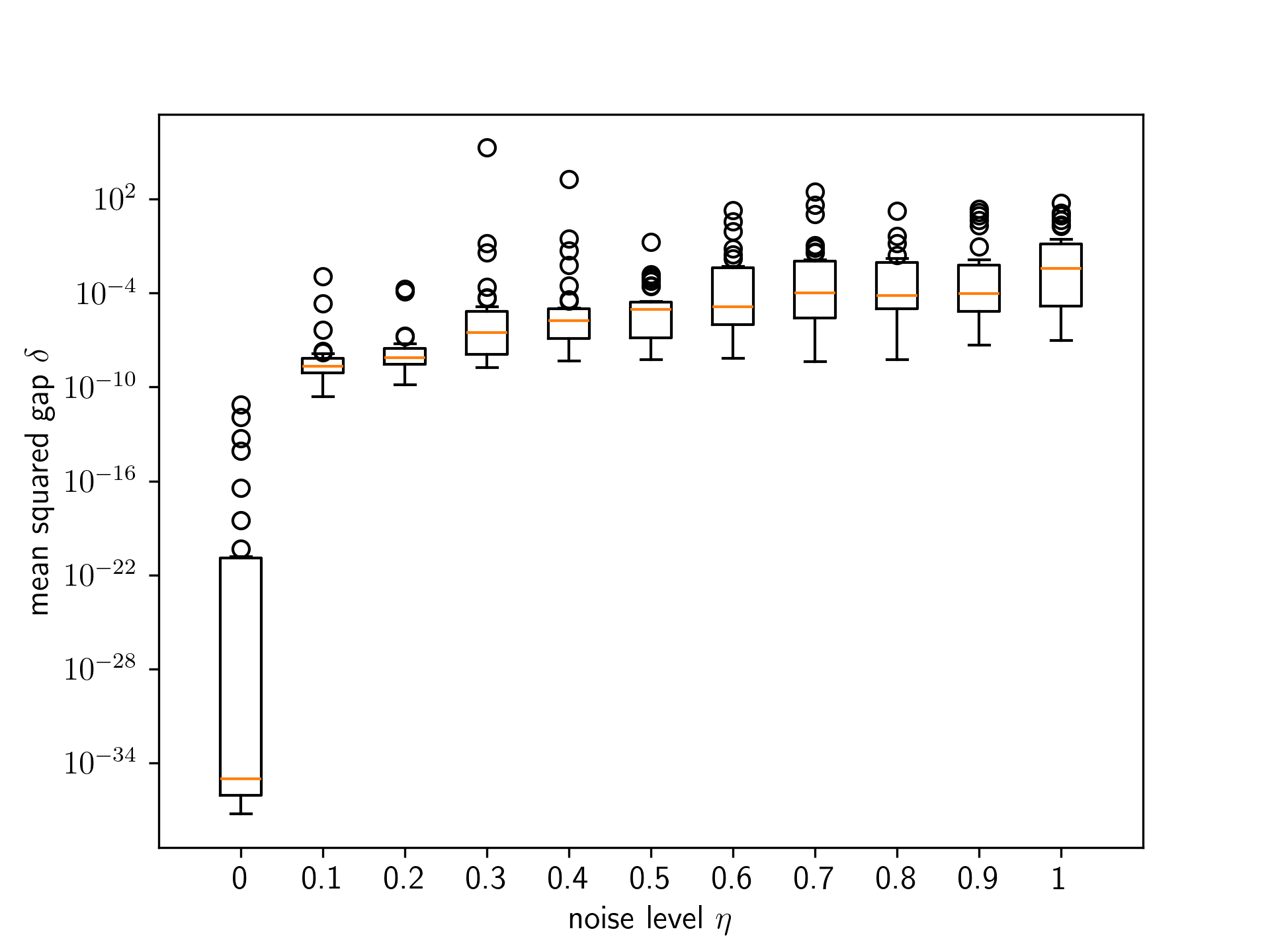}
    \hfill~
    \caption{Task III: results obtained when the noise level $\noiseLevel$ sweeps the $[0,1]$ interval. Left: box-and-whisker plots showing the distributions of $\euclideanError$. Right: box-and-whisker plots showing the distributions of $\meanSquareGap$. These plots show (1)~that we can recover the relationship between the target domain and the source domains, under the mixture assumption, when the target domain belongs to the \odd and (2)~that the $\meanSquareGap$ can be used to determine when the target domain belongs to the \odd.}
    \label{fig:task-III-results}
\end{figure}

\section{Conclusion}
\label{sec:conclusion}
In this work, we present a novel approach to characterize domains by estimating the distribution of physical parameters. 
Our method enhances interpretability, facilitates domain adaptation, and provides safeguards for systems operating outside their \odd.
Our experiments, organized around three tasks, are performed in the particular, but very important, case of autonomous vehicles and demonstrate how to obtain an absolute characterization of a domain by predicting a distribution of weather parameters (1)~using a single image acquired by a vehicle-mounted camera and (2)~using a bag of images. Our experiments also demonstrate (3)~how to obtain a relative characterization of a target domain based on arbitrarily chosen source domains.
Our approach includes two types of domain characterization: absolute and relative. The relative characterization is particularly valuable for domain adaptation, allowing the expression of a target domain in terms of source domains and verifying whether the current domain is part of the \odd. This is important for autonomous driving systems as well as for other fields requiring interpretable and trustworthy domain adaptation.

\begin{credits}
\subsubsection{\ackname} This work has been made possible thanks to the \emph{TRAIL} initiative (\url{https://trail.ac}). Part of it was supported by the Walloon region (Service Public de Wallonie Recherche, Belgium) under grant n°2010235 (ARIAC by DIGITALWALLONIA4.AI). J. Held and A. Cioppa are funded by the F.R.S.-FNRS (\url{https://www.frs-fnrs.be/en/}). 

\subsubsection{\discintname}
The authors have no competing interests to declare that are relevant to the content of this article.
\end{credits}

%
%

\end{document}